\documentclass{article}
\usepackage{arxiv}
\usepackage[utf8]{inputenc}
\usepackage[T1]{fontenc}
\usepackage{graphicx}
\usepackage{booktabs}
\usepackage{multirow}
\usepackage{amsmath,amssymb,amsfonts}
\usepackage{textcomp}
\usepackage{microtype}
\usepackage[numbers,sort&compress]{natbib}
\usepackage{url}
\usepackage{xcolor}
\usepackage{hyperref}
\hypersetup{colorlinks=true,linkcolor=blue!45!black,citecolor=blue!45!black,urlcolor=blue!45!black}
\providecommand{\tightlist}{\setlength{\itemsep}{0pt}\setlength{\parskip}{0pt}}

\renewcommand{\today}{}
\title{Measuring Curriculum Alignment across Topical Coverage, Competency, and Cognitive Depth: A Cross-Generational Framework Applied to CS2013 and CS2023}

\author{
  Sherzod Turaev\thanks{Corresponding author: \texttt{sherzod@uaeu.ac.ae}}\\
  Dept.\ of Computer Science and Software Engineering\\
  College of Computing and Artificial Intelligence\\
  United Arab Emirates University, Al Ain, UAE
  \and
  Saja Al-Dabet\\
  Dept.\ of Artificial Intelligence\\
  Aqaba University of Technology, Aqaba, Jordan
  \and
  Mary John\\
  Academic Support Department\\
  Abu Dhabi Polytechnic, Abu Dhabi, UAE
  \and
  Mamoun Awad\\
  Dept.\ of Computer Science and Software Engineering\\
  United Arab Emirates University, Al Ain, UAE
  \and
  Nazar Zaki\\
  Dept.\ of Computer Science and Software Engineering\\
  United Arab Emirates University, Al Ain, UAE
  \and
  Khaled Shuaib\\
  Dept.\ of Information Systems and Security\\
  United Arab Emirates University, Al Ain, UAE
}

\begin{document}
\maketitle

\begin{abstract}
Undergraduate computer science is governed by international curricular guidelines revised about once a decade, yet programs lack a reliable way to measure how completely they cover the current guideline and how coverage shifts when it changes. Existing analyses rely on topic models or manual tagging, seldom report reliability, do not benchmark the matching method, and examine topical overlap at a single point in time. We address these gaps with a staged pipeline that separates candidate generation from confirmation, applied to one accredited Bachelor of Science in Computer Science against Computer Science Curricula 2013 (CS2013) and 2023 (CS2023). Semantic retrieval proposes candidate course-to-knowledge-unit matches, a large language model confirms each against an explicit coverage rule, and an independent domain expert validates the resulting map. Benchmarking seven retrievers against pooled relevance judgments, we find that no automatic configuration reaches acceptable precision and recall together, peaking at an F1 of 0.55 and inflating apparent coverage past ninety percent once tuned for recall, which establishes retrieval as a candidate generator, not a measurement. Each map was validated by two independent domain experts and reconciled to a consensus, with substantial first-pass agreement (Cohen's kappa 0.64 and 0.69); the reported coverage is the lenient end of a threshold-sensitivity band whose strict end lies about seven points lower. Coverage of CS2023 is 48.4 percent of knowledge units, 59.4 percent by recommended hours, and about 28 percent of topics, and sixty-nine percent of covered units rest on a single course. The program articulates most competencies it covers yet meets the recommended cognitive depth far less often under CS2023 than under CS2013, a gap that survives a sensitivity analysis of the level mapping, while structural gaps stay separable from artifacts of the standard's evolution. The instrument is reusable and released.
\end{abstract}

\vspace{0.5em}

\vspace{0.5em}

\hypertarget{introduction}{%
\section{Introduction}\label{introduction}}

The undergraduate computer science curriculum is governed, more than in most disciplines, by a shared international reference. Since the first Curriculum 68 report, the Association for Computing Machinery and the IEEE Computer Society have periodically codified what an undergraduate degree in computer science should contain, expressing it as a \emph{body of knowledge} organized into knowledge areas, knowledge units, and topics with recommended hours and learning outcomes. The two most recent of these reports frame the present study. Computer Science Curricula 2013 organized the discipline into eighteen knowledge areas on a tiered core-and-elective model \citep{cs2013}, and its successor, Computer Science Curricula 2023, restructured the discipline into seventeen areas, elevated artificial intelligence and the social and professional dimensions of computing, and introduced a mathematical and statistical foundations area \citep{cs2023}. These guidelines are not merely advisory documents; they inform program design, they are the reference against which accreditation bodies such as the Computing Accreditation Commission of ABET frame their program criteria \citep{abet2025}, and they shape what graduates are expected to know. A program's degree of alignment with the current guideline is therefore a question of real consequence for curriculum committees, accreditors, and students alike.

Knowing where a concrete program stands against the guideline is, however, surprisingly difficult to establish rigorously. The guideline specifies what should be taught, while a program records what it teaches in a different vocabulary, namely, course learning outcomes and syllabus topics, so that measuring coverage requires a mapping between two large bodies of semi-structured text. In practice, this mapping is performed informally, through expert reading, which is laborious, hard to reproduce, and difficult to keep current. The difficulty is compounded by the fact that the reference itself moves. Because the guideline is revised roughly once a decade, a program designed under one generation of the standard must eventually be judged against the next, and the question is no longer only how completely a program covers the body of knowledge but how its coverage shifts when the body of knowledge is restructured. Neither the static nor the cross-generational question is answered by the guidelines themselves. Coverage, moreover, is only the first of the questions a program must answer. Establishing that a topic is taught does not establish that the corresponding \emph{competency} is articulated in a course learning outcome, nor that it is taught at the \emph{cognitive depth} the guideline recommends, and it is competencies and their depth, rather than topic lists alone, that the disciplinary guidelines and the accreditation criteria increasingly emphasize \citep{cc2020, frezza2018, abet2025}, a shift that the labor market appears to mirror \citep{turaev2026labor}. A complete account of alignment therefore requires three nested lenses of increasing stringency: whether the content is taught at all, whether the competency is articulated as a stated outcome, and whether that outcome reaches the recommended cognitive level.

A growing literature has begun to bring computation to bear on curriculum analysis, yet it leaves the measurement problem incompletely solved. One strand measures program alignment to the ACM guidelines directly, for example by projecting collections of syllabi onto the CS2013 knowledge-area space with topic models \citep{cs2013lda}, by tagging course materials to standards for coverage and audit \citep{csmaterials}, or by representing curricula through ontologies \citep{ontology}. A second strand automates the alignment of educational text with language technology, mapping course learning outcomes to program learning outcomes with reported precision against expert judgment \citep{cloplo} and classifying pedagogical materials against curriculum guidelines using word embeddings and large language models \citep{matclass}. These studies establish that coverage analysis is feasible and that embeddings and language models can align educational text. Though valuable, they share limitations that matter for the question at hand: they rely on topic models or manual tagging and seldom report the reliability of their judgments, they do not benchmark which retriever or model is appropriate for the task but adopt one by assumption, they evaluate model output against expert precision rather than treating it as a candidate to be confirmed, they map to institutional outcomes or to course materials rather than to the external disciplinary body of knowledge, and they are confined to a single snapshot of a single guideline. Decisively for competency-based accreditation, they also measure topical overlap alone, without asking whether a program's stated outcomes articulate the guideline's competencies or deliver them at the recommended cognitive depth. To overcome these drawbacks, a method is needed that measures coverage against the external standard, that is benchmarked and reliability-checked rather than assumed, that confirms candidates under an explicit rule and checks that confirmation against expert judgment rather than trusting automated similarity to constitute the result, that extends beyond topical coverage to competency articulation and cognitive depth, and that can be applied across guideline generations.

In the current research, we develop such a method and apply it across the two guideline generations to an accredited Bachelor of Science in Computer Science, mapping the same program against both CS2013 and CS2023 through the three nested lenses of topical coverage, competency articulation, and cognitive depth. The approach pairs semantic retrieval, which proposes candidate matches between the program's text and the guideline's knowledge units and learning outcomes, with confirmation by a large language model governed by explicit definitions of coverage, competency delivery, and depth-adequacy, and it validates both the retriever and the resulting maps empirically. The contributions are the following:

\begin{enumerate}
\def\labelenumi{\arabic{enumi}.}
\tightlist
\item
  a reproducible instrument that measures a program's coverage of an external computing body of knowledge, pairing a benchmarked, high-recall retriever with confirmation under an explicit rule, performed by a large language model and validated by independent experts with chance-corrected reliability;
\item
  a quantification of the ceiling on automation, namely that no fully automatic configuration attains acceptable precision and recall together and that a rule based on embedding similarity alone over-identifies competency delivery about twofold, which bears directly on a literature that evaluates model output against expert precision and then reports that output as the measurement;
\item
  a three-lens reading of the same program, from topical coverage, through the articulation of competencies in stated outcomes with an explicit \emph{articulation gap}, to the cognitive depth at which they are delivered, which reveals a depth gap that is small against CS2013 but substantial against CS2023;
\item
  an accreditation-grounded diagnosis that reads the coverage gaps against the ABET program criteria and separates deliberate specialization from accreditation-relevant omission; and
\item
  a cross-generational comparison across all three lenses that separates gaps persisting across a decade from changes that are artifacts of the standard's own evolution.
\end{enumerate}

The paper is organized as follows. The next section reviews related work on the curricular guidelines, on curriculum mapping and coverage analysis, and on the language-technology methods on which we build. The Materials and Methods section then describes the program and the two bodies of knowledge as structured corpora and details the nine stages of the pipeline, which extend from topical coverage to competency articulation and cognitive depth. The Results section reports the retriever benchmark and reliability checks, the coverage of the current standard, the accreditation-grounded gap diagnosis, the competency presence and cognitive-depth findings, and the cross-generational comparison. The Discussion interprets these findings for programs that must track an evolving standard and for accreditation practice, and considers the limits of automation. The final section concludes and identifies future work, principally the extension of the method across multiple institutions.

\hypertarget{related-work}{%
\section{Related Work}\label{related-work}}

\hypertarget{curricular-guidelines-and-their-evolution}{%
\subsection{Curricular guidelines and their evolution}\label{curricular-guidelines-and-their-evolution}}

The ACM and IEEE Computer Society have issued curricular guidelines for computer science at roughly decade intervals, and the transition this study exploits is the most recent. The lineage runs from Computing Curricula 2001 \citep{cc2001} through the cross-disciplinary Computing Curricula 2020 competency framework \citep{cc2020} to the two reports that frame this study. Computer Science Curricula 2013 organized the body of knowledge into eighteen knowledge areas, with topics labeled Core-Tier1, Core-Tier2, or Elective, and with recommended hours and learning outcomes for each unit \citep{cs2013, sahami2013}. Computer Science Curricula 2023 retained the area-unit-topic structure but reorganized the discipline into seventeen areas under a CS-Core, KA-Core, and Non-core classification, expanded the treatment of artificial intelligence and of society, ethics, and the profession, and consolidated the mathematical prerequisites into a dedicated mathematical and statistical foundations area \citep{cs2023, raj2023cs2023}. The guidelines define what should be taught and at what depth, but they neither prescribe a method for measuring how completely a given program covers them nor address how a program's coverage shifts as the guideline is revised. Those two questions motivate the present work.

\hypertarget{curriculum-mapping-and-coverage-analysis}{%
\subsection{Curriculum mapping and coverage analysis}\label{curriculum-mapping-and-coverage-analysis}}

A body of research measures how programs align with curricular guidelines. Closest to ours in aim, a syllabus-level study projected computer science departments onto the CS2013 knowledge-area space using a simplified supervised topic model, comparing coverage across institutions and observing that programs emphasize different areas \citep{cs2013lda}. At the level of whole programs, \citet{blumenthal2022} measured five hundred accredited computer science programs against the CS2013 body of knowledge and the ABET criteria, the breadth counterpart to the unit-level, cross-generational reading we pursue here. Tool-based approaches such as CS Materials allow educators to tag course materials to standards for the purposes of design, alignment, audit, and search \citep{csmaterials}, and related efforts pursue the visual understanding of curricula \citep{visualcurricula}, ontological representations of curricula and learning material \citep{ontology}, and systematic reviews of semantic technologies in the computer science curriculum \citep{semreview}. This work establishes coverage analysis as an active genre, yet it rests on topic models or manual tagging, rarely reports the reliability of its mappings, does not connect the resulting gaps to accreditation criteria, and examines a single point in time. The current research contributes a model-confirmed, expert-validated map at the level of knowledge units, an accreditation-grounded reading of the gaps, and a comparison across two guideline generations.

\hypertarget{automated-alignment-of-outcomes-and-materials}{%
\subsection{Automated alignment of outcomes and materials}\label{automated-alignment-of-outcomes-and-materials}}

A second line of work applies natural-language processing and large language models to the alignment of educational text. Most relevant, a system automated the mapping of course learning outcomes to program learning outcomes, reporting precision against domain experts for two programs and arguing that such automation can support program evaluation \citep{cloplo}. More recent systems apply large language models to map syllabi to competency frameworks \citep{kalish2026} and to evaluate model performance on curricular-analytics tasks \citep{xu2025}, and text embeddings have been used to align educational content standards across frameworks \citep{butterfuss2025}. Others classify pedagogical materials against computer science curriculum guidelines using both classical word-embedding methods and pre-trained large language models \citep{matclass}, and study the progression of educational topics through semantic matching \citep{topicprog}. In our own prior work, we survey the use of large language models for aligning computing curricula with the external labor market and set out a reference framework and design desiderata for such systems \citep{turaev2026labor}. Broader reviews document the rapid adoption of natural-language processing across educational tasks, from feedback analysis to assessment \citep{shaik2022, kastrati2021}. These studies demonstrate that embeddings and language models can align educational text at useful accuracy. They differ from the present work in four respects that bear on the measurement problem: they evaluate against expert precision rather than chance-corrected reliability, they do not benchmark which retriever or model suits the task, they treat the model output as the result rather than as candidates for rule-based confirmation, and they target institutional outcomes, course materials, or the labor market rather than an external disciplinary body of knowledge tracked over time.

\hypertarget{methods-and-instruments}{%
\subsection{Methods and instruments}\label{methods-and-instruments}}

The pipeline composes established techniques. Dense sentence embeddings build on transformer language models \citep{devlin2019} and follow Sentence-BERT \citep{sbert} and the embedding families we benchmark, namely BGE-M3 \citep{bgem3}, E5 \citep{e5}, and GTE \citep{gte}. The retrieval framing and the use of cosine similarity in a vector space follow standard information-retrieval practice \citep{manning2008}, within which this task sits between lexical scoring \citep{robertson2009} and dense neural retrieval \citep{karpukhin2020}, while the benchmarks now used to compare embedding models motivate measuring the retriever on the task rather than importing it from a general leaderboard \citep{thakur2021, muennighoff2023}. Combining the outputs of several retrievers uses reciprocal rank fusion \citep{rrf}, and constructing the evaluation reference by pooling the candidates of all systems follows established relevance-judgment practice \citep{voorhees2000}. Inter-rater agreement is reported with Cohen's kappa \citep{cohen1960}, interpreted on the scale of \citet{landiskoch} and read in the light of later guidance on the statistic \citep{mchugh2012}, agreement of this text-coding kind having a well-established basis in computational linguistics \citep{artstein2008}. The educational scaffolding draws on constructive alignment \citep{biggs1996, biggs}, on outcome-based education \citep{spady1994} and on curriculum mapping as a method for making a program's coverage transparent \citep{harden2001, uchiyama2008}, on Bloom's taxonomy \citep{bloom1956} in its revised form \citep{anderson2001, krathwohl2002} and as adapted for computing \citep{ccecc, fuller2007, thompson2008}, on the competency paradigm articulated for the computing disciplines \citep{cc2020, frezza2018}, and on the ABET computing program criteria \citep{abet2025} against which the gaps are read. To our knowledge, no prior curriculum-coverage study assembles these elements into a benchmarked, agreement-validated, cross-generational instrument.

In sum, prior work either describes the guidelines without measuring program coverage, measures coverage by topic models or manual tagging without reliability or accreditation grounding, or aligns outcomes and materials with language technology against expert precision on a single snapshot, and in every case it stops at topical overlap, leaving the articulation of competencies and their cognitive depth unmeasured. The current research differs on five axes at once: it benchmarks the retriever rather than assuming one, it treats retrieval as candidate generation for rule-based confirmation and reports chance-corrected agreement on the resulting maps, it grounds the diagnosed gaps in accreditation criteria, it extends the analysis from topical coverage to competency articulation and cognitive depth on a Bloom scale adapted for computing \citep{ccecc}, and it measures the same program against two guideline generations.

\hypertarget{materials-and-methods}{%
\section{Materials and Methods}\label{materials-and-methods}}

\hypertarget{study-design}{%
\subsection{Study design}\label{study-design}}

\begin{figure}[!ht]\centering
\includegraphics[width=1\linewidth]{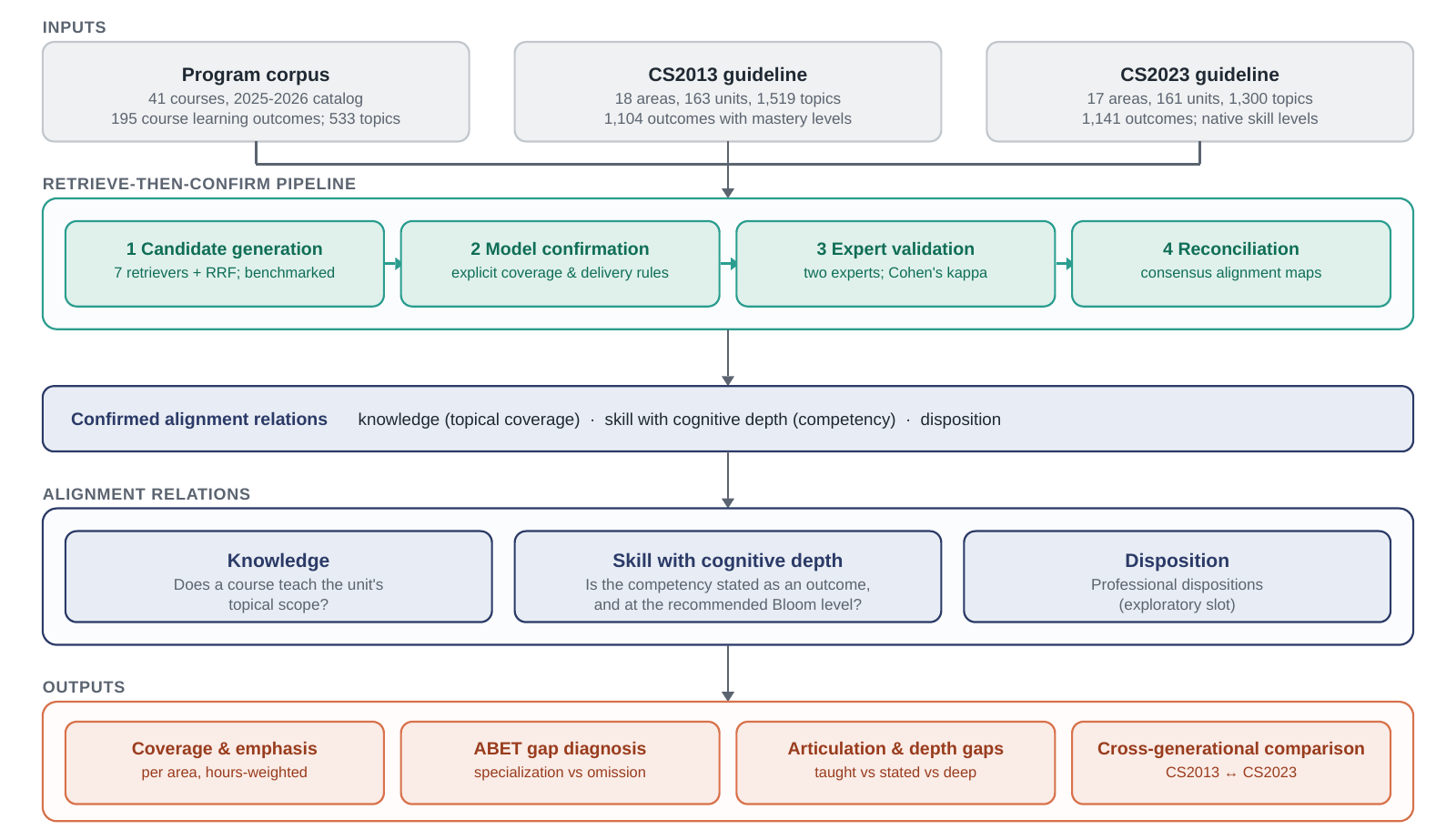}
\caption{Overview of the curriculum-alignment framework. The program and the two guideline corpora feed a benchmarked retrieve-then-confirm pipeline, whose confirmed output is expressed as three alignment relations, namely knowledge, skill with cognitive depth, and disposition, and read through the reported analyses; the same procedure is applied to both CS2013 and CS2023. The four pipeline steps shown are applied across the nine analysis stages described in the Methods, seven topical and two for competency}\label{fig:framework}
\end{figure}

The study measures how completely one undergraduate computer science program covers the disciplinary body of knowledge defined by the ACM/IEEE/AAAI curricular guidelines, and how that coverage differs when the program is read against the previous guideline, CS2013 \citep{cs2013}, and the current one, CS2023 \citep{cs2023}. Because a guideline expresses its content as a hierarchy of knowledge areas, knowledge units, and topics, while a program expresses its content as course learning outcomes and syllabus topics, the central operation is a mapping between two bodies of semi-structured text that no single automatic method performs with sufficient reliability on its own. We therefore designed a staged pipeline in which semantic retrieval proposes candidate matches, a large language model confirms them against an explicit definition of coverage, an independent expert rater establishes the reliability of those confirmations, and the reconciled map is then read for coverage, emphasis, accreditation-relevant gaps, competency articulation, cognitive depth, and cross-generational change. The pipeline comprises nine stages, described in turn below; the first seven measure and interpret topical coverage, and the last two extend the same retrieve-then-confirm design from topics to the articulation of competencies and to their cognitive depth. Throughout, the design separates the cheap, high-recall act of generating candidates from the expensive, high-precision act of confirming them, so that automation accelerates the work without being trusted to constitute the result. Figure~\ref{fig:framework} gives an overview of the inputs, the retrieve-then-confirm pipeline, the three alignment relations it confirms, and the analyses they support.

The program studied is the Bachelor of Science in Computer Science offered by the Department of Computer Science and Software Engineering at the College of Computing and Artificial Intelligence, United Arab Emirates University, in the 2025-2026 catalog. The program is accredited and is the subject of recurring self-study against both the ACM/IEEE guidelines and the criteria of the Computing Accreditation Commission of ABET \citep{abet2025}, which makes it a representative and consequential case for a coverage analysis. The program's most recent accreditation was completed in 2023, before the CS2023 final report was endorsed and released in 2024 \citep{cs2023}, so the curriculum was last reviewed under the expectations of the previous guideline generation; this timing makes the question of its alignment to the current standard both concrete and timely, and it is precisely the situation that the cross-generational analysis is designed to address.

\hypertarget{materials}{%
\subsection{Materials}\label{materials}}

\hypertarget{the-program-corpus}{%
\subsubsection{The program corpus}\label{the-program-corpus}}

The program side of the analysis consists of 41 catalog courses spanning the general-education, college-requirement, major-requirement, internship, and elective components of the degree. From the official course descriptions and syllabi we extracted, for each course, its catalog learning outcomes and its syllabus topics, together with course metadata, namely the course code, title, credit hours, and curricular block. The extraction yielded 195 catalog learning outcomes and 533 syllabus topics, for a total of 728 program text items that serve as the units to be matched. Syllabus topics were retained in their teaching order as listed in the source documents rather than alphabetized, since the order carries pedagogical information that aids judgment during confirmation. Every extracted record carries the path and a cryptographic hash of its source file, so that each item is traceable to the exact document state from which it was drawn.

\hypertarget{the-two-bodies-of-knowledge-as-structured-corpora}{%
\subsubsection{The two bodies of knowledge as structured corpora}\label{the-two-bodies-of-knowledge-as-structured-corpora}}

The reference side consists of the two curricular guidelines, each converted from its published form into a structured corpus with the same schema. For each guideline we recorded its knowledge areas, the knowledge units within each area, the topics within each unit, the tier or core classification of each topic, the recommended hour allocations, and the illustrative learning outcomes. CS2023 was represented by 17 knowledge areas, 161 knowledge units, 1,300 topics classified as CS Core (422), KA Core (538), and Non-core (340), and 1,141 illustrative learning outcomes. CS2013 was represented by 18 knowledge areas, 163 knowledge units, 1,519 topics classified as Core-Tier1 (306), Core-Tier2 (357), and Elective (856), and 1,104 learning outcomes annotated with their mastery level, namely Familiarity (582), Usage (370), and Assessment (152). The CS2013 corpus was extracted programmatically from the 518-page report and validated against the document's own summary table: the parsed Tier-1 and Tier-2 core-hour totals matched the official figures for all eighteen knowledge areas, which provides external confirmation that the extraction preserved the source structure faithfully. The learning outcomes of both guidelines, together with the program's 195 catalog learning outcomes, form the basis of the competency analysis described below, and the mastery annotations of CS2013 and the native skill levels of CS2023 supply the recommended cognitive level against which the depth of the program's outcomes is judged. As with the program corpus, each record retains source provenance, including the verified hash of the source document.

To make the two guidelines comparable, we authored a crosswalk that aligns each CS2013 knowledge area with its CS2023 counterpart. Seventeen of the eighteen CS2013 areas correspond to a CS2023 area, in most cases through renaming, for example Intelligent Systems becoming Artificial Intelligence, Information Management becoming Data Management, Programming Languages becoming Foundations of Programming Languages, and Social Issues and Professional Practice becoming Society, Ethics, and the Profession; Computational Science is the single area without a CS2023 successor. The crosswalk is released as an artifact so that the cross-generational alignment is fully auditable.

\hypertarget{a-formal-framework-for-the-three-alignment-relations}{%
\subsection{A formal framework for the three alignment relations}\label{a-formal-framework-for-the-three-alignment-relations}}

Having described the program and the two guidelines as structured corpora, we formalize the alignment as three relations between a program and a guideline, which makes the measurement precise and its aggregation reproducible. A terminological distinction is needed here and is maintained throughout. The three \emph{alignment relations} defined in this subsection, namely knowledge, skill with cognitive depth, and disposition, follow the CS2023 decomposition of a competency and describe what is measured about a unit. The three \emph{lenses} named in the Introduction, namely topical coverage, competency articulation, and cognitive depth, are the successive readings of increasing stringency that the study reports, and they correspond to the knowledge relation and to the two components of the skill relation. The disposition relation is defined for completeness of the decomposition and is instantiated only as an exploratory lower bound, for the reason given in the Results, so it supports no headline finding. The formalism follows the CS2023 conception of a \emph{competency} as knowledge applied through skills under appropriate dispositions, so it measures one relation per component, with the cognitive-depth dimension carried inside the skill relation; the nine stages described below instantiate these relations, the first seven computing the knowledge relation and the last two the skill relation.

A program is a set of courses $C=\{c_1,\dots,c_n\}$, each course $c$ supplying a set of course learning outcomes $O(c)$, its stable competency commitments, and a set of syllabus topics $T_P(c)$, the more volatile content of a particular offering; write $\Omega=\bigcup_c O(c)$ for all program outcomes. A guideline is organized into knowledge areas $\mathcal{A}$ and knowledge units $\mathcal{U}$, with $\mathrm{ka}(u)\in\mathcal{A}$ the area of unit $u$, and each unit $u$ carries a topic set $T(u)$ and a learning-outcome set $L(u)$. Cognitive depth is the ordinal scale
\[ \Lambda=\{1<2<3<4<5\}, \]
read as Remember, Understand, Apply, Analyze/Evaluate, and Create, onto which the native schemes map: the CS2013 mastery levels Familiarity, Usage, and Assessment to $2$, $3$, and $4$, and the CS2023 unit skill levels Explain, Apply, and Develop or Evaluate to $2$, $3$, and $4$ or $5$. Each guideline outcome carries a recommended level $\lambda^{*}(\ell)\in\Lambda$, and each program outcome a delivered level $\lambda(o)\in\Lambda$ obtained from its leading action verb through the Bloom's-for-Computing taxonomy.

The first relation, \emph{knowledge}, is topical coverage: a unit is covered when some course substantively teaches its topical scope,
\[ \mathrm{cov}_K(u)=\mathbf{1}\!\left[\,\exists\,c:\ c\ \text{substantively teaches}\ T(u)\,\right]. \]
The second relation, \emph{skill with depth}, takes the outcome rather than the topic as the unit of analysis. Let $M\subseteq\Omega\times\bigcup_u L(u)$ be the confirmed competency-matching relation, where $(o,\ell)\in M$ means the program outcome $o$ delivers the competency expressed by the guideline outcome $\ell$. An outcome delivers a unit when it delivers any of that unit's outcomes, so that, writing the delivering set $\Delta(u)=\{o\in\Omega:\exists\,\ell\in L(u),\ (o,\ell)\in M\}$, a covered unit is competency-\emph{present} and \emph{depth-adequate} respectively when
\[ \mathrm{cov}_S(u)=\mathbf{1}\!\left[\,\Delta(u)\neq\varnothing\,\right],\qquad
   \mathrm{cov}_S^{\Lambda}(u)=\mathbf{1}\!\left[\,\Delta(u)\neq\varnothing\ \wedge\ \max_{o\in\Delta(u)}\lambda(o)\ \ge\ \lambda^{*}(u)\,\right], \]
where $\lambda^{*}(u)$ is the unit's recommended level, taken as the median of the native levels of the unit's outcomes. Aggregating over the units a program covers in an area $a$ yields a presence rate $S(a)$ and a depth-adequacy rate $S^{\Lambda}(a)$, and the difference $S(a)-S^{\Lambda}(a)$ isolates \emph{under-depth} coverage. The third relation, \emph{disposition}, matches program outcomes to the professional dispositions $D(a)$ that CS2023 lists for each area through $M_D\subseteq\Omega\times\bigcup_a D(a)$, with area coverage $D\text{-cov}(a)=|\{\delta\in D(a):\exists\,o,\ (o,\delta)\in M_D\}|/|D(a)|$; because dispositions are coarse and only weakly observable from stated outcomes, this slot is defined here for completeness and treated as exploratory rather than as a headline result.

Separating the three relations is what lets the analysis distinguish three failure modes that a single topical figure conflates: content that is absent ($\mathrm{cov}_K=0$); a competency that is absent although its content is present ($\mathrm{cov}_K=1$, $\mathrm{cov}_S=0$), which we term the \emph{articulation gap}; and a competency that is present but taught below the recommended level ($\mathrm{cov}_S=1$, $\mathrm{cov}_S^{\Lambda}=0$), which we term the \emph{depth gap}. The knowledge relation is computed from topics and is therefore sensitive to syllabus drift, whereas the skill relation is computed from the stated outcomes that accreditation and the labor market hold a program to, which is the reason the competency lens complements rather than duplicates the topical one.

\hypertarget{stage-1-candidate-generation-by-semantic-retrieval}{%
\subsection{Stage 1: Candidate generation by semantic retrieval}\label{stage-1-candidate-generation-by-semantic-retrieval}}

The first stage proposes, for each course, a shortlist of knowledge units it may cover, framed as an information-retrieval problem in which the course's text items are queries and the knowledge units are the documents to be retrieved. To avoid the truncation that long concatenated unit descriptions would impose on models with short context windows, and to keep every comparison between short, comparable pieces of text, we represented each knowledge unit by its constituent topics together with its name, and we scored a course against a unit as the maximum cosine similarity over all pairs formed by the course's items and the unit's topics. This maximum-aggregation choice rewards a unit whose core content is matched by any part of the course rather than requiring the whole unit to resemble the whole course.

We computed similarity with seven retrievers so that the choice of retriever could be made empirically rather than assumed. Six were neural sentence-embedding models, namely BGE-M3 \citep{bgem3}, BGE-large-en-v1.5, E5-large-v2 \citep{e5}, GTE-large \citep{gte}, all-mpnet-base-v2, and all-MiniLM-L6-v2 \citep{sbert}, each applied with its recommended query and passage formatting and with embeddings L2-normalized so that the inner product equals the cosine similarity. The seventh was a term-frequency-inverse-document-frequency baseline over word unigrams and bigrams, included to quantify how much semantic embedding adds over lexical overlap on this task.

\hypertarget{stage-2-retriever-benchmarking-with-pooled-relevance-judgments}{%
\subsection{Stage 2: Retriever benchmarking with pooled relevance judgments}\label{stage-2-retriever-benchmarking-with-pooled-relevance-judgments}}

To select a retriever without biasing the choice toward any one system, we evaluated the seven retrievers against a reference set built by pooling, following standard information-retrieval evaluation practice \citep{voorhees2000}. For each course we formed the union of the top ten candidate units proposed by every retriever, and the confirmation stage rated this pooled set under the coverage rule, so that no retriever's blind spots could distort the ground truth in its own favor. Each retriever was then scored against the pooled, confirmed reference using recall at several cut-offs, mean reciprocal rank, mean average precision, normalized discounted cumulative gain at ten, and the shortlist depth required to reach ninety-five percent mean recall, the last of these expressing the practical confirmation burden each retriever imposes. We also evaluated a reciprocal-rank-fusion ensemble that combines the seven ranked lists \citep{rrf}, using the standard fusion constant. The ensemble was adopted for candidate generation on the strength of this benchmark, and its margin over the strongest single model, together with the unexpectedly weak performance of a reputedly strong long-context model on this short-text task, is reported in the Results.

\hypertarget{stage-3-human-confirmation-of-the-coverage-map}{%
\subsection{Stage 3: Model confirmation of the coverage map}\label{stage-3-human-confirmation-of-the-coverage-map}}

Retrieval produces plausible candidates, not coverage, so the third stage converts the shortlist into a map through confirmation under an explicit rule. The confirmation was performed by a large language model prompted with each course's outcomes and topics, the candidate unit, and the coverage rule stated below, and these model confirmations were then validated by an independent expert rater, as Stage 4 describes. A course was recorded as covering a knowledge unit when it \emph{substantively} covers it, which we defined operationally as teaching the core of the unit's scope as material a student would learn and be assessed on, rather than sharing only a keyword, assuming the unit as a prerequisite, or mentioning it in passing. Coverage was judged against the unit's scope, that is its constituent topics, rather than against its title; partial but genuine treatment of a unit's core was counted as coverage, since the complementary question of how completely each unit is covered is addressed separately at the topic level. Each confirmed match was annotated with a confidence level, high for substantive coverage and medium for lighter or partial treatment, and with a provenance tag recording whether the unit had been surfaced by retrieval or recovered by the rater beyond the retrieved shortlist. The same rule and annotations were applied to both guidelines. The candidate pool for both guidelines was the multi-retriever pool of Stage 2, so that candidate generation was identical across the two standards. An earlier version of the CS2013 analysis had instead combined lexical retrieval with crosswalk seeding, whereby every CS2013 unit lying in an area that corresponds to a CS2023 area the course already covers was added to the course's candidate set; we replaced that procedure with the benchmarked ensemble to remove any dependence of the previous-standard map on the current-standard map, and we report the effect of the change in the Results.

\hypertarget{stage-4-inter-rater-reliability-and-reconciliation}{%
\subsection{Stage 4: Inter-rater reliability and reconciliation}\label{stage-4-inter-rater-reliability-and-reconciliation}}

To establish that the model's confirmations are not merely an automated artifact, an independent expert rater, a senior computing faculty member with long involvement in the program's accreditation who took no part in the model's confirmation pass, judged a blinded sample of candidate pairs, and agreement between the model and the expert was quantified with Cohen's kappa \citep{cohen1960} following standard procedure for inter-rater reliability \citep{hallgren2012, krippendorff2004} and interpreted on the scale of \citet{landiskoch}. Because each judgment is a binary, nominal coverage decision by two raters, the unweighted coefficient is the appropriate one; the weighted \citep{cohen1968} and multi-rater \citep{fleiss1971} variants were not required here. For CS2023 the sample was a balanced set of 274 course-to-unit candidates, half of which the model had marked as covered and half not, presented without the model's labels. For CS2013 the second-rater pass was focused on the units that required judgment beyond mirroring the CS2023 map, namely the areas where the two guidelines diverge in granularity or structure, comprising 127 blinded candidates. Disagreements were then reconciled by a documented rule rather than ad hoc: matches the model had confirmed with high confidence were held, matches confirmed with only medium confidence on which the expert disagreed were deferred to the stricter verdict, the generic per-area society-ethics-and-profession sub-units were excluded to avoid double-counting professional content captured by the dedicated area, and a small number of clearly substantive additions proposed by the expert were accepted. Reconciliation produced the consensus maps used in all subsequent analyses, and it additionally surfaced one systematic correction to the CS2013 crosswalk, namely that a teamwork outcome is not equivalent to the Software Project Management unit, which was retained only where genuine project management is taught. Because a coverage judgment turns on how much of a unit's core a course must teach to count, and the model and a single expert can reasonably draw that line differently, we strengthened the validation beyond the balanced sample in two ways. The first expert reviewed the entire confirmed positive map rather than a sample, and a second independent expert, also a senior member of the program's faculty and likewise uninvolved in the model's confirmation pass, then adjudicated every course-to-unit pair on which the model and the first expert disagreed, grading each by how much of the unit's core the course teaches on a three-level scale of core, partial, or absent. A strict reading of coverage counts only the units graded as fully taught, whereas the lenient reading, which is the one the coverage rule states, counts partial but genuine treatment as well. The map reported throughout is the reconciliation at the lenient reading, and every coverage figure is accompanied by the strict figure as a sensitivity bound, so that the dependence of the result on where the threshold is drawn is made explicit rather than hidden inside a single number.

\hypertarget{stage-5-coverage-emphasis-and-topic-level-estimation}{%
\subsection{Stage 5: Coverage, emphasis, and topic-level estimation}\label{stage-5-coverage-emphasis-and-topic-level-estimation}}

From each consensus map we computed coverage as the number of distinct knowledge units that at least one course covers, reported overall, per knowledge area, and per core tier, and additionally weighted by the guideline's recommended hours so that areas the guideline emphasizes count proportionally. We characterized the program's emphasis by attributing each course's credit hours across the areas it covers and comparing the resulting program share against the guideline's recommended-hour share, the difference indicating where the program weights an area more or less heavily than the guideline advises. We also recorded, for each covered unit, the courses that cover it, which exposes how concentrated or dispersed curricular responsibility is and how many units depend on a single course.

Coverage at the finer granularity of individual topics was estimated rather than exhaustively confirmed, because confirming every one of the more than one thousand topics per guideline by hand is infeasible and, as we found, the similarity signal at that granularity is intrinsically weak: within a unit the program already covers, every topic is close in meaning to the course, so similarity separates taught from untaught sub-topics only modestly. We therefore estimated topic coverage with the strongest single retriever from the benchmark and calibrated its decision threshold against a sample of forty topics that a rater adjudicated by hand, selecting the threshold that maximized agreement and reporting the resulting precision and recall so that the estimate is qualified by its measured accuracy rather than presented as exact.

To assess whether the confirmation stage could be dispensed with, we measured how closely the fully automatic ensemble, at a range of fixed shortlist depths, reproduces the confirmed consensus map, reporting the precision, recall, and F1 of the automatic course-to-unit pairs against the consensus.

\hypertarget{stage-6-gap-analysis-against-accreditation-criteria}{%
\subsection{Stage 6: Gap analysis against accreditation criteria}\label{stage-6-gap-analysis-against-accreditation-criteria}}

To distinguish deliberate specialization from consequential omission, we read the coverage gaps against the program criteria for computer science of the Computing Accreditation Commission of ABET \citep{abet2025}, which require substantial coverage of certain areas, namely algorithms and complexity, computer science theory, the concepts of programming languages, and software development, and exposure to others, namely architecture and organization, information management (the data-management area in CS2023 terms), networking and communication, operating systems, and parallel and distributed computing. Mapping the program's measured coverage onto these requirements classifies each gap as an accreditation-relevant shortfall or as an acceptable emphasis consistent with the program's mission.

\hypertarget{stage-7-cross-generational-comparison-across-guideline-generations}{%
\subsection{Stage 7: Cross-generational comparison across the two guidelines}\label{stage-7-cross-generational-comparison-across-guideline-generations}}

The cross-generational analysis applies the entire pipeline a second time, against CS2013, and then aligns the two consensus maps through the area crosswalk so that coverage can be compared area by area across the two guideline generations. The comparison yields a program-relative difference between the standards, which we read for three patterns: gaps that persist across both guidelines and are therefore structural rather than recent, areas whose apparent coverage changes only because the guideline was restructured, and content that one guideline recognizes and the other does not.

\hypertarget{stage-8-competency-articulation-of-learning-outcomes}{%
\subsection{Stage 8: Competency articulation of learning outcomes}\label{stage-8-competency-articulation-of-learning-outcomes}}

Topical coverage establishes that a unit's content is taught; it does not establish that the corresponding competency is articulated in what the program promises its students, namely its course learning outcomes. Because competencies, rather than topic lists, are what accreditation and the labor market reward, and because the syllabus topics that constitute coverage may change from one offering to the next while the stated outcomes are comparatively stable, the eighth stage measures competency articulation directly. Following the CS2023 and CC2020 conception of a competency as knowledge applied through skills under appropriate dispositions \citep{cc2020, frezza2018}, we operationalized the skill dimension as a relation between the program's 195 course learning outcomes and the guidelines' learning outcomes, in contrast to the topical stages, which matched courses to knowledge units through their topics.

The competency-matching relation $M$ of the framework was built with the same benchmarked ensemble used for topical retrieval. For each covered unit, every course learning outcome was scored against the unit as its maximum similarity to the unit's learning outcomes, and the highest-scoring outcomes of the covering courses were pooled as candidates for confirmation. A unit was recorded as \emph{competency-present}, that is $\mathrm{cov}_S(u)=1$, when its delivering set $\Delta(u)$ is non-empty, which holds when at least one outcome of a covering course genuinely delivers the unit's competency, defined operationally as the outcome's stated ability matching at least one of the unit's own learning outcomes rather than sharing only a subject area or naming a different ability the same course also teaches. Presence is therefore \emph{outcome-based}: a unit whose content is taught but whose competency is not stated in any outcome, that is a unit with $\mathrm{cov}_K(u)=1$ and $\mathrm{cov}_S(u)=0$, is recorded as not present and is reported separately as the \emph{articulation gap}, the set of units the program teaches yet does not articulate. The fifteen dispositional outcomes, which express professional behaviors such as functioning on teams and communicating, carry no cognitive verb and were routed to the disposition slot rather than the skill relation.

As in the topical stages, similarity proposes candidates, a large language model confirms them under the delivery rule, and an independent expert validates the result, and confirmation proved indispensable here for a measurable reason reported in the Results: the automated similarity-with-threshold rule agreed with the expert rater on the matching relation at only a slight-to-fair level, labeling roughly twice as many outcome-to-unit pairs as matches as the confirmed relation contains, so the relation cannot be read off similarity. Reliability was therefore established with two blinded second-rater exercises, one on a stratified sample of outcome-to-unit candidate pairs and one on unit-level presence, each rated independently and quantified with Cohen's kappa.

\hypertarget{stage-9-cognitive-depth-and-depth-adequacy}{%
\subsection{Stage 9: Cognitive depth and depth-adequacy}\label{stage-9-cognitive-depth-and-depth-adequacy}}

A competency may be articulated yet pitched below the level the guideline expects, so the ninth stage measures the cognitive depth of the articulated competencies. Every learning outcome, on both the program and the guideline side, was placed on a five-level scale derived from Bloom's taxonomy \citep{bloom1956} in its revised form \citep{anderson2001, krathwohl2002} as adapted for computing \citep{ccecc, thompson2008}: Remember as level one, Understand as level two, Apply as level three, Analyze and Evaluate together as level four, and Create as level five. The level of each program outcome was assigned from its leading action verb through a verb-to-level lexicon built from the computing taxonomy, an approach related to the automated classification of assessment items by Bloom level \citep{jayakodi2016, mohammed2020}, and confirmed by hand for the outcomes whose verb was ambiguous, absent, or non-cognitive. To establish the reliability of this program-outcome level coding, an independent rater re-classified all 195 program outcomes on the same five-level scale, blinded to the automated assignment, and agreement was almost perfect, at ninety-three percent exact and ninety-six percent within one level, with a quadratically weighted Cohen's $\kappa$ of 0.924 and a $\kappa$ of 0.972 for the distinction between a cognitive and a dispositional outcome. The recommended level of each unit was taken from native information wherever the guideline supplies it, namely the unit skill levels of CS2023 and the Familiarity, Usage, and Assessment mastery annotations of CS2013, mapped to levels two, three, and four respectively, and from verb classification of the unit's outcomes otherwise; because CS2013 supplies a human-assigned mastery level for every outcome, it served as a built-in test of the verb classifier, which agreed exactly with the native level in sixty-three percent of cases and within one level in eighty-four percent, an accuracy that justifies preferring native levels where they exist.

A present unit was judged \emph{depth-adequate}, that is $\mathrm{cov}_S^{\Lambda}(u)=1$, when $\max_{o\in\Delta(u)}\lambda(o)\ge\lambda^{*}(u)$, that is when, among the outcomes that genuinely deliver it, the highest delivered cognitive level meets or exceeds the unit's recommended level; the level is read only from the delivering outcomes in $\Delta(u)$, never from a higher-level outcome the same course supplies for a different unit, a distinction that proved decisive for reliability. We report the area presence rate $S$ and depth-adequacy rate $S^{\Lambda}$ of the framework, the under-depth gap $S-S^{\Lambda}$, and, as the cleanest expression of cognitive alignment, the \emph{depth-among-delivered} rate $S^{\Lambda}/S$, namely the fraction of competency-present units that are also depth-adequate. The depth judgment was validated by a third blinded second-rater exercise over the present units, again quantified with Cohen's kappa under the delivering-outcome rule.

\hypertarget{implementation-reproducibility-and-use-of-artificial-intelligence}{%
\subsection{Implementation, reproducibility, and use of artificial intelligence}\label{implementation-reproducibility-and-use-of-artificial-intelligence}}

The pipeline was implemented in Python. Lexical similarity used the scikit-learn implementation of term-frequency-inverse-document-frequency, and the neural embeddings were computed with the sentence-transformers library; the candidate scores, similarity matrices, and ranked lists were cached so that every reported figure is reproducible from fixed intermediate outputs. All extracted records carry source paths and SHA-256 hashes, and the structured corpora, the consensus maps for both guidelines, the reconciliation logs, the area crosswalk, the retriever-benchmark scripts, and the rater tools are released as supplementary material.

Two of the pipeline's stages use generative artificial intelligence as part of the method itself rather than as a writing aid, and we disclose them here in full. The confirmation stages of the pipeline, both the topical course-to-unit confirmation of Stage 3 and the competency confirmation of Stage 8, were performed by a large language model prompted with the relevant program text, the candidate guideline item, and the explicit coverage or delivery rule; these model confirmations constitute the first-pass map, which an independent expert rater then validated, as described in Stages 4, 8, and 9. Corpus extraction from the two guideline reports likewise used large language models, documented with the released extraction code. The models, their versions, and the prompts are released with the artifacts so that every model-assisted step can be inspected and repeated. No generative system is credited as an author, and the authors take responsibility for the accuracy of all content, including every citation.

\hypertarget{results}{%
\section{Results}\label{results}}

\hypertarget{retriever-benchmark-and-method-validation}{%
\subsection{Retriever benchmark and method validation}\label{retriever-benchmark-and-method-validation}}

\begin{figure}[!ht]\centering
\includegraphics[width=1\linewidth]{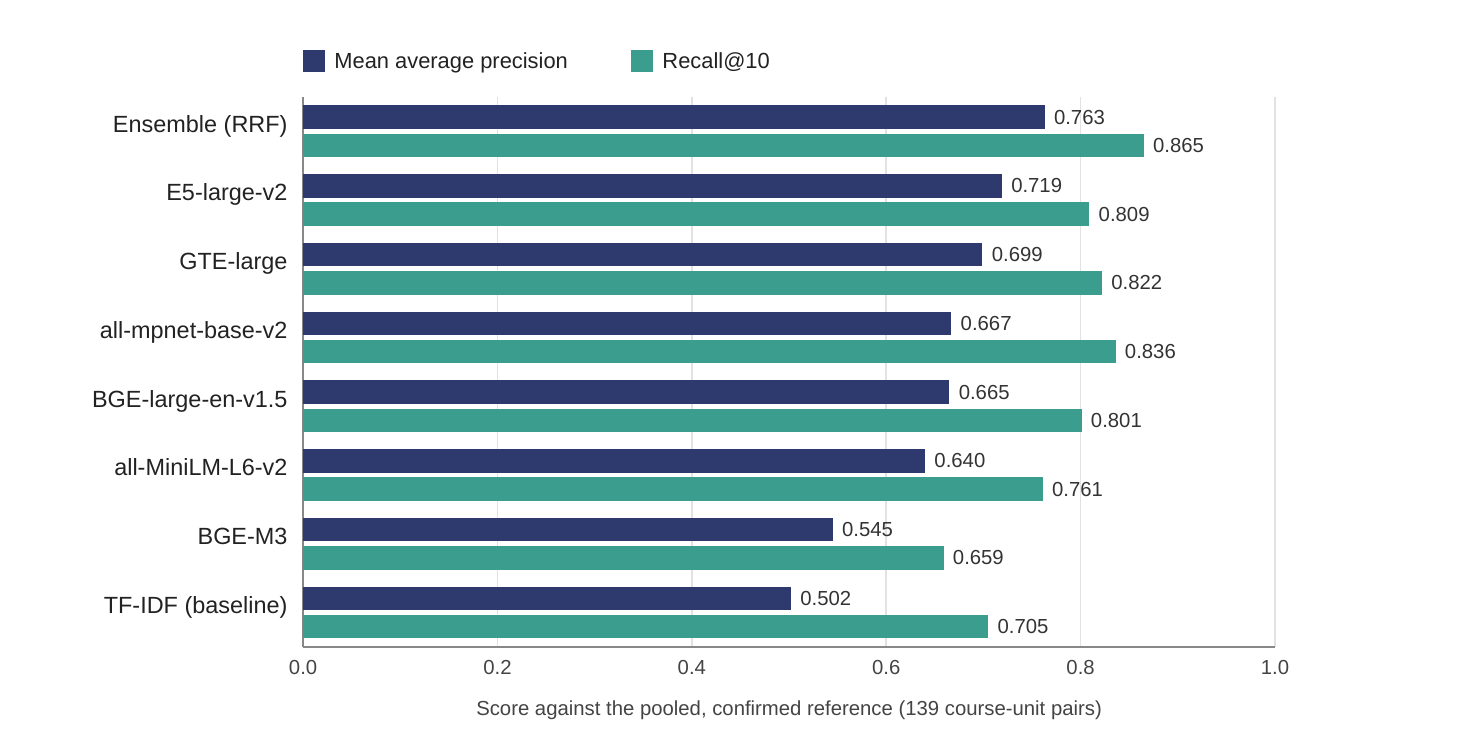}
\caption{Retriever benchmark: mean average precision and recall at ten against the pooled, confirmed reference}\label{fig:leaderboard}
\end{figure}

The seven retrievers differed substantially on the pooled, confirmed reference of 139 course-to-knowledge-unit pairs across 36 courses (Table~\ref{tab:leaderboard}; Figure~\ref{fig:leaderboard}). The reciprocal-rank-fusion ensemble was the strongest configuration on every metric, reaching a mean average precision of 0.763 and recovering ninety-five percent of true units within a shortlist of eighteen candidates, against thirty for the strongest single model. Among the single models the ranking was led by E5-large-v2 (mean average precision 0.719) and GTE-large (0.699), with the lightweight all-mpnet-base-v2 close behind (0.667) and notably efficient, reaching ninety-five percent recall at a shortlist of twenty despite its smaller size. The term-frequency-inverse-document-frequency baseline sat at the bottom on mean average precision (0.502), which confirms that lexical overlap alone is the limiting configuration and that semantic embedding contributes real gain on this task.

\begin{table}[!ht]\centering\small\renewcommand{\arraystretch}{1.3}\setlength{\tabcolsep}{7pt}
\caption{Retriever benchmark against the pooled, confirmed reference (139 course--knowledge-unit pairs, 36 courses). Best value per column in bold. $k_{95}$ is the shortlist depth for 95\% mean recall}\label{tab:leaderboard}
\setlength{\tabcolsep}{4pt}
\begin{tabular}{lccccccc}\toprule
Retriever & R@5 & R@10 & R@20 & MRR & MAP & nDCG@10 & $k_{95}$\\ \midrule
Ensemble (RRF) & \textbf{0.714} & \textbf{0.865} & \textbf{0.967} & \textbf{0.931} & \textbf{0.763} & \textbf{0.825} & \textbf{18}\\
E5-large-v2 & 0.664 & 0.809 & 0.888 & 0.867 & 0.719 & 0.772 & 30\\
GTE-large & 0.654 & 0.822 & 0.907 & 0.880 & 0.699 & 0.771 & 28\\
all-mpnet-base-v2 & 0.708 & 0.836 & 0.950 & 0.852 & 0.667 & 0.753 & 20\\
BGE-large-en-v1.5 & 0.649 & 0.801 & 0.903 & 0.844 & 0.665 & 0.738 & 24\\
all-MiniLM-L6-v2 & 0.615 & 0.761 & 0.921 & 0.872 & 0.640 & 0.714 & 29\\
BGE-M3 & 0.544 & 0.659 & 0.760 & 0.780 & 0.545 & 0.620 & --\\
TF-IDF (baseline) & 0.549 & 0.705 & 0.828 & 0.706 & 0.502 & 0.598 & --\\
\bottomrule\end{tabular}\end{table}

One result runs against expectation and is worth stating plainly. BGE-M3, a model with a strong general reputation, finished second from last (mean average precision 0.545), below the much smaller all-MiniLM-L6-v2 and only marginally above the lexical baseline. The pattern is consistent with a task effect rather than a defect of the model: the matching here is between very short English phrases under maximum-similarity aggregation, a regime that favors models tuned for sentence-level similarity over a model engineered for long-context, multilingual retrieval whose distinctive sparse and multi-vector capabilities are not exercised by the uniform dense interface. The finding is the empirical justification for benchmarking the retriever on the actual task rather than importing a choice from general leaderboards.

The confirmed maps proved reliable against expert judgment. On CS2023, the model's confirmations and the expert rater's independent verdicts over a blinded, balanced sample of 274 candidates agreed on 224 of them, for a raw agreement of 0.818 and a Cohen's kappa of 0.635, which is substantial on the scale of Landis and Koch. On CS2013, a focused pass over the 127 candidates that required judgment beyond mirroring the CS2023 map gave a raw agreement of 0.843 and a Cohen's kappa of 0.685, again substantial (Table~\ref{tab:reliability}). The disagreements were not random but concentrated where the model had recorded low confidence and on the generic per-area society-ethics-and-profession sub-units. Of the fifty CS2023 disagreements, the thirty-four the model had marked covered and the expert had not split evenly by the model's recorded confidence, seventeen high-confidence matches being held and seventeen medium-confidence matches deferred to the stricter verdict, while of the sixteen the expert read more broadly, five generic society-ethics-and-profession sub-units were excluded to avoid double-counting and three substantive additions were accepted. The net effect changed the CS2023 map by eight units, from 88 to 80 covered units, and the CS2013 map by one unit, from 84 to 83, without disturbing the area-level pattern. Of the 127 course-to-unit pairs in the confirmed CS2023 map before the final reconciliation, 124 were drawn from the retriever shortlist and only three were recovered by the rater beyond it, so the map rests on the pooled candidates rather than on manual backfilling. A fuller expert review sharpened this picture, and the disagreement it surfaced is instructive. When the first expert reviewed the entire confirmed positive map rather than the balanced sample, the expert upheld 86 of the 124 shortlist-derived CS2023 course-to-unit pairs and 76 of the 123 CS2013 pairs, a stricter reading than the model's that concentrated on units the model had credited from partial or adjacent treatment rather than from the unit's core. The second expert then adjudicated the 89 pairs on which the model and the first expert disagreed. Read strictly, counting only fully taught cores, the second expert's grades agree with the first expert on 75 percent of those pairs; read leniently, counting partial treatment as the coverage rule does, they agree with the model on 89 percent. The three judgments therefore concur on what the courses teach and differ only on where to place the coverage threshold. Reconciling at the lenient reading gives the coverage reported here, 78 of 161 CS2023 units (48.4 percent) and 86 of 163 CS2013 units (52.8 percent), which moves the balanced-sample maps of 80 and 83 covered units to their reconciled totals, while the strict reading gives 41.6 and 45.4 percent; the strict figures bound the sensitivity of every coverage result to the threshold, and the near-equality of the program's alignment across the two guidelines holds under both readings.

\begin{table}[!ht]\centering\small\renewcommand{\arraystretch}{1.3}\setlength{\tabcolsep}{7pt}
\caption{Inter-rater reliability of the coverage maps. Confusion entries are counts of (model, expert) verdicts}\label{tab:reliability}
\begin{tabular}{lccccccc}\toprule
Guideline & Items & Raw agr. & Cohen's $\kappa$ & cover/cover & cover/not & not/cover & not/not\\ \midrule
CS2023 & 274 & 0.818 & 0.635 & 103 & 34 & 16 & 121\\
CS2013 (focused) & 127 & 0.843 & 0.685 & 53 & 11 & 9 & 54\\
\bottomrule\end{tabular}\end{table}

Automation alone did not reproduce the confirmed map. The fully automatic ensemble peaked at a pair-level F1 of 0.55, at a shortlist depth of five, where it recovered course-to-unit pairs at a precision of 0.48 and a recall of 0.63; deeper shortlists raised recall toward 0.95 only by admitting so many false positives that precision fell below 0.20 and the apparent coverage inflated past ninety percent of all units. No depth achieved acceptable precision and recall together, which establishes that a confirmation stage reading each course against the unit's scope is necessary to the validity of the map and that retrieval, however well tuned, functions as a candidate generator rather than a coverage estimator. This ceiling is a property of similarity used as a classifier, and it is distinct from the confirmation stage, in which a language model reads the course and the unit under the coverage rule and whose confirmations agree with an expert at substantial kappa, as reported above. The point generalizes beyond this study. A literature that evaluates model output against expert precision on a favorable operating point, and then reports that output as the measurement, is reporting a quantity whose error we can now bound: at the recall the same literature would need in order to claim near-complete coverage, precision on this task falls below 0.20 and the apparent coverage inflates past ninety percent of all units, which is roughly twice the confirmed figure. The competency layer reproduces the pattern independently, where a rule based on embedding similarity alone agreed with the expert rater at a Cohen's kappa of only 0.30 and over-identified delivery about twofold. We therefore treat the ceiling on automation as a finding in its own right rather than as a preliminary to the coverage results.

\hypertarget{coverage-of-the-current-standard-cs2023}{%
\subsection{Coverage of the current standard (CS2023)}\label{coverage-of-the-current-standard-cs2023}}

\begin{figure}[!ht]\centering
\includegraphics[width=1\linewidth]{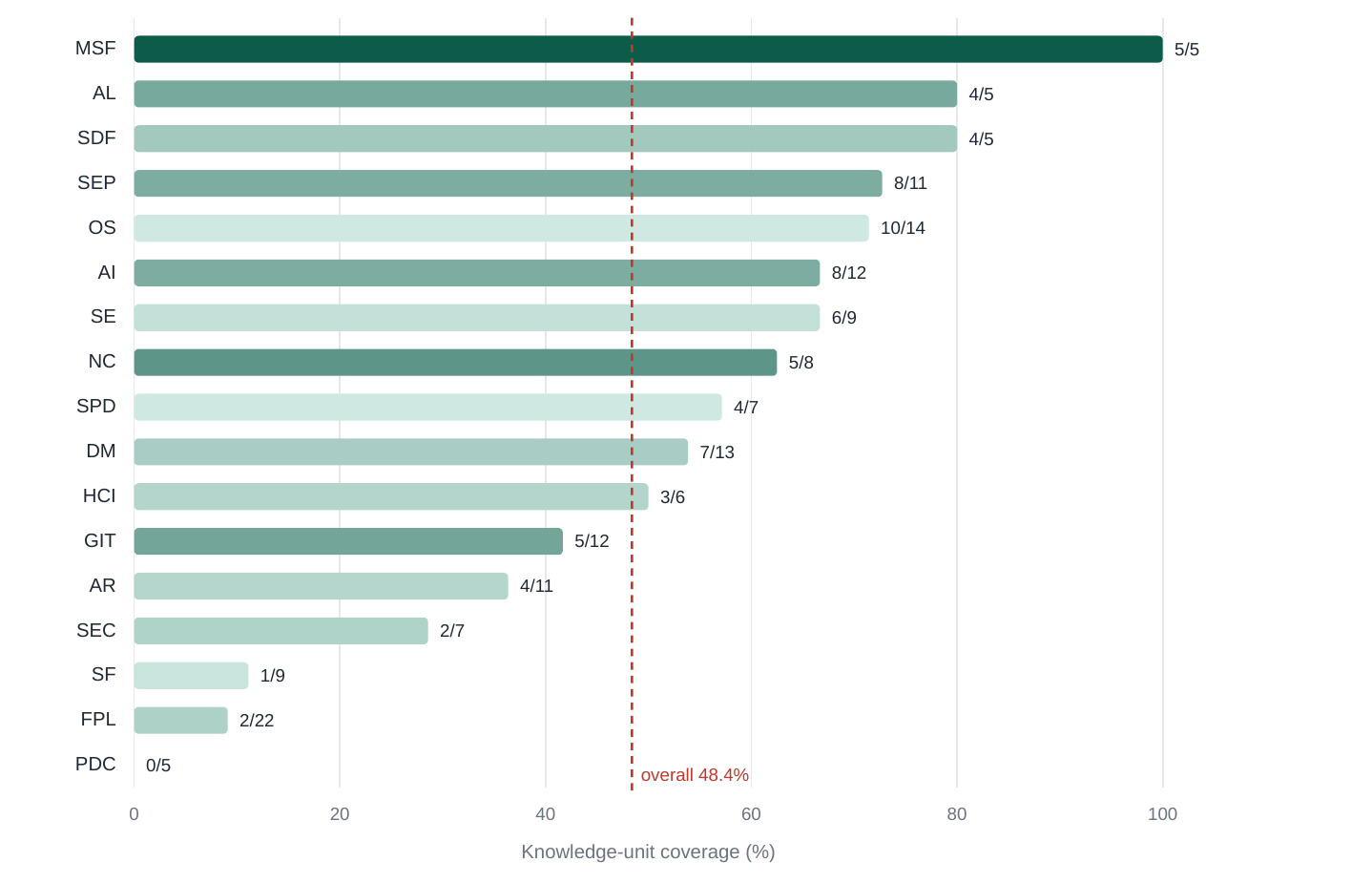}
\caption{Program coverage of CS2023 knowledge units by knowledge area; bar shade encodes the CS2023 recommended hours}\label{fig:coverage}
\end{figure}

The knowledge relation of the framework holds for 78 of the 161 CS2023 knowledge units, so that the program covers 48.4 percent of them ($\mathrm{cov}_K(u)=1$ for 78 units; Figure~\ref{fig:coverage}; Table~\ref{tab:cov23}). Because the well-covered areas tend to be those CS2023 weights most heavily, coverage weighted by the guideline's recommended hours is higher, at 59.4 percent, so the program is somewhat stronger when judged against where the discipline concentrates its recommended effort than the raw unit count suggests.

\begin{table}[!ht]\centering\small\renewcommand{\arraystretch}{1.3}\setlength{\tabcolsep}{7pt}
\caption{Program coverage of CS2023 knowledge units by knowledge area, with CS2023 recommended hours}\label{tab:cov23}
\begin{tabular}{lccc}\toprule
Knowledge area & Covered/Total & \% & Rec. hours\\ \midrule
Math \& Statistical Foundations & 5/5 & 100.0 & 120\\
Algorithmic Foundations & 4/5 & 80.0 & 67\\
Software Dev. Fundamentals & 4/5 & 80.0 & 45\\
Artificial Intelligence & 8/12 & 66.7 & 64\\
Society, Ethics, Profession & 8/11 & 72.7 & 64\\
Operating Systems & 10/14 & 71.4 & 23\\
Software Engineering & 6/9 & 66.7 & 29\\
Networking \& Comm. & 5/8 & 62.5 & 80\\
Specialized Platform Dev. & 4/7 & 57.1 & 23\\
Data Management & 7/13 & 53.8 & 42\\
Human-Computer Interaction & 3/6 & 50.0 & 37\\
Architecture \& Organization & 4/11 & 36.4 & 36\\
Graphics \& Interactive Tech. & 5/12 & 41.7 & 69\\
Security & 2/7 & 28.6 & 39\\
Systems Fundamentals & 1/9 & 11.1 & 26\\
Foundations of Programming Languages & 2/22 & 9.1 & 40.5\\
Parallel \& Distributed Comp. & 0/5 & 0.0 & 35\\
\midrule
All areas & 78/161 & 48.4 & 839.5\\
\bottomrule\end{tabular}\end{table}

Coverage is markedly uneven across knowledge areas. The program fully covers the mathematical and statistical foundations (5 of 5 units) and is strong in algorithmic foundations (4 of 5), software development fundamentals (4 of 5), artificial intelligence (8 of 12), society, ethics, and the profession (8 of 11), operating systems (10 of 14), and software engineering (6 of 9). At the other extreme it covers none of the five units of parallel and distributed computing, a single unit of nine in systems fundamentals, and two of twenty-two units in the foundations of programming languages, with security covered at only two of seven units. The intermediate band comprises networking (5 of 8), specialized platform development (4 of 7), data management (7 of 13), human-computer interaction (3 of 6), architecture and organization (4 of 11), and graphics and interactive techniques (5 of 12).

At the finer granularity of individual topics, coverage is lower and, as expected, harder to resolve. Calibrated against a hand-adjudicated sample, the topic-level estimator operated at a precision of 0.83 and a recall of 0.96, and it placed corrected topic coverage at approximately 28 percent of the full body of knowledge, with a clean monotonic ordering by tier in which the program reaches the essential CS-Core topics most fully and the Non-core topics least (about 41, 33, and 22 percent respectively at the calibrated operating point, before the precision-and-recall correction). The convergence of this estimate with an independent lexical estimate near the same value indicates that the topic figure is a validated approximation rather than an artifact of one estimator, and the divergence between it and the unit figure deserves to be stated rather than reconciled. Three defensible scalars describe the same program, namely 48.4 percent of knowledge units, 59.4 percent of recommended hours, and approximately 28 percent of topics, and they differ because they count different things: a unit is credited when a course substantively teaches its topical scope, which does not require that every topic within the unit is taught. The unit figure is therefore an upper bound read at the granularity at which the guideline organizes the discipline and at which a curriculum committee acts, and the topic figure a lower bound read at the granularity at which the guideline enumerates content. Neither is the coverage of the program in any absolute sense, and we report all three deliberately, since a single headline percentage invites precisely the over-reading that a reusable instrument should discourage. The product of the method is the map, and the percentages are summaries of it.

\begin{figure}[!ht]\centering
\includegraphics[width=1\linewidth]{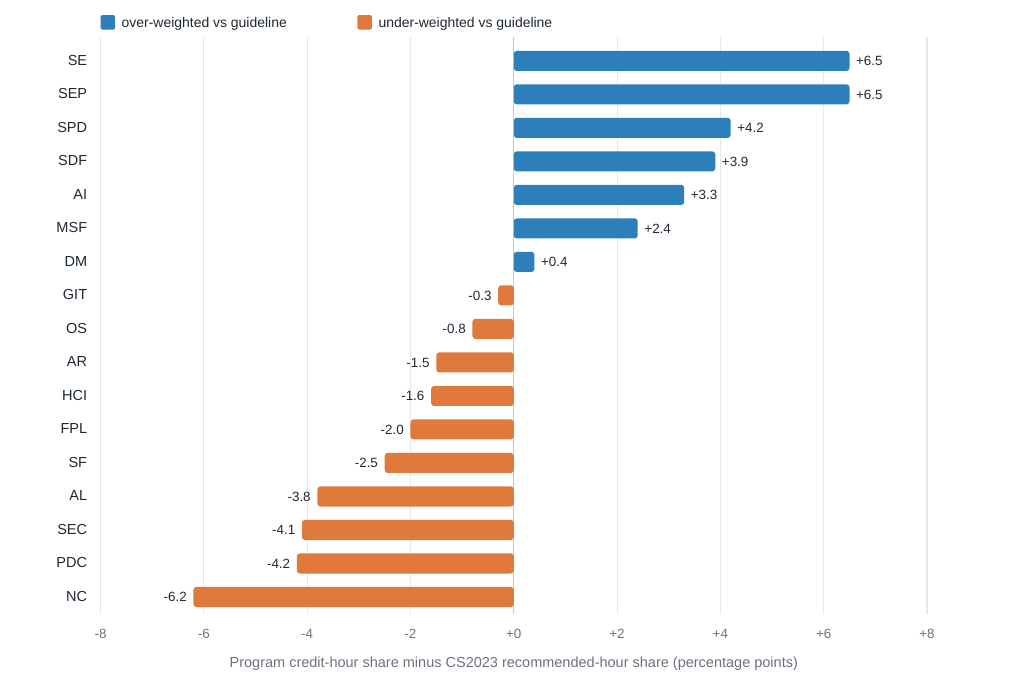}
\caption{Program emphasis minus CS2023 recommended-hour share, by knowledge area. Bars extending to the right of zero indicate over-weighting relative to the guideline and bars extending to the left indicate under-weighting, with the signed difference in percentage points labeled on every bar}\label{fig:emphasis}
\end{figure}

The program's emphasis departs from the guideline's recommended hours in a coherent way (Figure~\ref{fig:emphasis}). Measuring emphasis as each course's credit hours distributed across the areas it covers, the program over-weights software engineering and the society, ethics, and profession area most strongly, by about six percentage points of credit-hour share each, with specialized-platform development, software-development fundamentals, and artificial intelligence also above the recommended share; it under-weights networking by a comparable margin, followed by parallel and distributed computing, security, and algorithmic foundations, the last reflecting the guideline's heavy recommended-hour allocation to an area the program teaches largely through outcomes shared with other areas. This profile is consistent with a deliberately application-facing program rather than with uniform coverage. It is worth noting that the emphasis reading and the coverage reading are computed from different quantities, the first from credit-hour shares and the second from confirmed unit matches, and that they nevertheless converge on the same diagnosis: the areas the emphasis analysis flags as most under-weighted, namely networking, parallel and distributed computing, and security, are the areas in which the coverage analysis independently records the deepest shortfalls, at five of eight, none of five, and two of seven units respectively. The agreement of two independent measurements on the same three areas is evidence that the instrument is registering a property of the curriculum rather than an artifact of either construction.

Coverage rests on a small number of courses and is, in places, fragile, which is the most directly actionable result the analysis yields. Of the 78 covered units, 54, that is 69 percent, are covered by exactly one course, so retiring or substantially revising any one of those courses would remove the program's only contact with those units. A coverage percentage that appears stable at the level of the program is therefore single-threaded through individual courses across more than two thirds of its extent, and this is a property that no aggregate figure, and no accreditation self-study written at the level of program outcomes, would expose. The heaviest carriers are the operating-systems course with ten units and the second capstone, the artificial-intelligence course, the professional-responsibility course, the security course, and the software-engineering course with six to seven units each. Five courses map to no CS2023 unit, namely the three natural-science requirements, basic biology, and the bioinformatics elective, which is expected since their content lies outside the computer science body of knowledge.

\hypertarget{gap-diagnosis-against-accreditation-criteria}{%
\subsection{Gap diagnosis against accreditation criteria}\label{gap-diagnosis-against-accreditation-criteria}}

Reading the gaps against the ABET program criteria for computer science separates accreditation-relevant omissions from acceptable specialization (Table~\ref{tab:abet}). The criteria require substantial coverage of algorithms and complexity, computer science theory, the concepts of programming languages, and software development, and the program meets the first, second, and fourth of these comfortably while the third, the concepts of programming languages, is addressed for accreditation through the required programming sequence, even though the broader CS2023 foundations-of-programming-languages area is sparse, covered at only two of twenty-two units, with no dedicated programming-languages or compilers course. The criteria also require exposure to architecture and organization, information management, networking and communication, operating systems, and parallel and distributed computing; the program meets all of these except parallel and distributed computing, which it does not cover at the unit level and which is therefore the second criterion-relevant omission. The lighter treatment of systems fundamentals, the single-course treatment of security, and the under-weighting of networking relative to recommended hours are consistent with the program's emphasis and do not conflict with the criteria, since the criteria require only exposure in those areas and that bar is met.

\begin{table}[!ht]\centering\small\renewcommand{\arraystretch}{1.3}\setlength{\tabcolsep}{7pt}
\caption{Diagnosed gaps read against the ABET computing program criteria for computer science. The criteria's ``information management'' requirement is read against the CS2023 data-management area}\label{tab:abet}
\begin{tabular}{p{4.6cm}p{2.6cm}p{2.2cm}p{2.6cm}}\toprule
CS2023 area & ABET requirement & Coverage & Verdict\\ \midrule
Algorithmic Foundations & substantial & 80\% & met\\
Computer science theory (AL-Models) & substantial & covered & met\\
Foundations of Programming Languages & substantial & 9\% & sparse at unit level\\
Software Dev. Fundamentals / Software Engineering & substantial & 80\% / 67\% & met\\
Architecture \& Organization & exposure & 36\% & met\\
Data Management & exposure & 54\% & met\\
Networking \& Communication & exposure & 62\% & met (under-weighted)\\
Operating Systems & exposure & 71\% & met\\
Parallel \& Distributed Computing & exposure & 0\% & \textbf{unmet (omission)}\\
\bottomrule\end{tabular}\end{table}

\hypertarget{cross-generational-comparison-cs2013-against-cs2023}{%
\subsection{Cross-generational comparison: CS2013 against CS2023}\label{cross-generational-comparison-cs2013-against-cs2023}}

\begin{figure}[!ht]\centering
\includegraphics[width=1\linewidth]{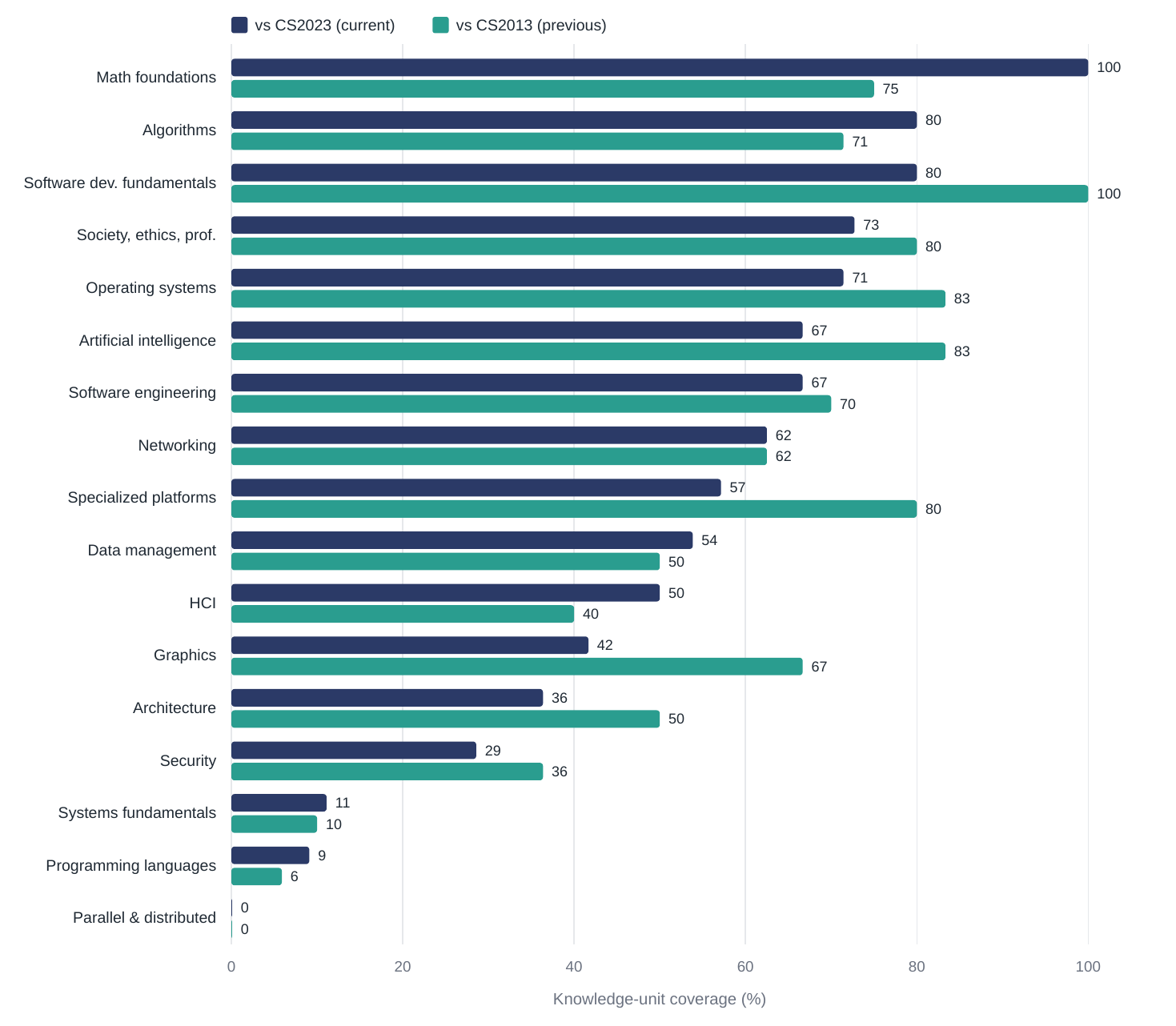}
\caption{Coverage of the same program against CS2013 and CS2023 by aligned knowledge area}\label{fig:comparison}
\end{figure}

Applying the pipeline a second time, against CS2013, the program covers 86 of 163 units, that is 52.8 percent, close to its 48.4 percent against CS2023 (Figure~\ref{fig:comparison}; Table~\ref{tab:cmp}). The near-equality of the totals is itself a finding: the program's degree of alignment to the guideline has remained essentially constant across a decade of disciplinary revision. Because the two candidate pools were generated by the same benchmarked ensemble, this near-equality is not an artifact of how candidates were produced; when the CS2013 candidates are generated by the ensemble rather than by the earlier crosswalk-seeded procedure, the ensemble recalls 86 percent of the confirmed pairs directly, the remainder are recovered by the rater, and the additional units the ensemble proposes are, on adjudication under the same rule, either cross-area false positives or units the CS2023 map declines for the same course, so the near-equality reflects the program and the two standards rather than the way candidates were generated. What changes is the composition rather than the amount.

\begin{table}[!ht]\centering\small\renewcommand{\arraystretch}{1.3}\setlength{\tabcolsep}{7pt}
\caption{Program coverage against CS2013 and CS2023 by aligned knowledge area (knowledge-unit level). The Math foundations row combines the CS2013 Discrete Structures and Computational Science areas, which the crosswalk aligns with the CS2023 mathematical and statistical foundations area for comparability}\label{tab:cmp}
\begin{tabular}{lll}\toprule
Aligned area & CS2013 & CS2023\\ \midrule
Algorithms & 5/7 (71\%) & 4/5 (80\%)\\
Architecture & 4/8 (50\%) & 4/11 (36\%)\\
Math foundations & 9/12 (75\%) & 5/5 (100\%)\\
Graphics & 4/6 (67\%) & 5/12 (42\%)\\
HCI & 4/10 (40\%) & 3/6 (50\%)\\
Security & 4/11 (36\%) & 2/7 (29\%)\\
Data management & 6/12 (50\%) & 7/13 (54\%)\\
Artificial intelligence & 10/12 (83\%) & 8/12 (67\%)\\
Networking & 5/8 (62\%) & 5/8 (62\%)\\
Operating systems & 10/12 (83\%) & 10/14 (71\%)\\
Specialized platforms & 4/5 (80\%) & 4/7 (57\%)\\
Parallel \& distributed & 0/9 (0\%) & 0/5 (0\%)\\
Programming languages & 1/17 (6\%) & 2/22 (9\%)\\
Software dev. fundamentals & 4/4 (100\%) & 4/5 (80\%)\\
Software engineering & 7/10 (70\%) & 6/9 (67\%)\\
Systems fundamentals & 1/10 (10\%) & 1/9 (11\%)\\
Society, ethics, prof. & 8/10 (80\%) & 8/11 (73\%)\\
\midrule
Overall & 86/163 (52.8\%) & 78/161 (48.4\%)\\
\bottomrule\end{tabular}\end{table}

Three patterns emerge from the aligned-area comparison. First, several gaps persist across both guidelines and are therefore structural rather than recent. Parallel and distributed computing is covered at zero percent against both, and it is important that this area existed in CS2013, so its absence is a decade-long omission rather than a failure to adopt a new area; the same holds, in weaker form, for the foundations of programming languages (six percent then, nine percent now) and systems fundamentals (zero then, eleven percent now). Second, some apparent change is driven by the evolution of the guideline rather than the program. CS2013 confined mathematics to discrete structures and a small computational-science area, with no units for calculus, linear algebra, or statistics, whereas CS2023 introduced the mathematical and statistical foundations area that includes them. The consequence is best stated at the level of courses rather than units. The program's discrete-mathematics content had CS2013 units to occupy, which is why the aligned mathematics row still reads nine of twelve under CS2013, but its calculus, linear-algebra, and statistics courses had no CS2013 unit to map onto at all and were therefore invisible to a CS2013 reading, whereas under CS2023 they populate the new area completely, at five of five. The revision in effect recognizes a part of the program's mathematics core that the older standard gave no place to record. The mirror image appears in computational science, an area CS2013 contained and CS2023 removed, with which the program's modeling-and-simulation elective aligns, a small signature of a curriculum still partly anchored to the older guideline. Third, restructuring within an area changes the picture without any change in what the program teaches, most visibly in security, where the program's single course covers a higher fraction of CS2013's eleven-unit area (36 percent) than of CS2023's reorganized seven-unit area (29 percent).

\hypertarget{competency-presence-and-cognitive-depth}{%
\subsection{Competency presence and cognitive depth}\label{competency-presence-and-cognitive-depth}}

Moving from topical coverage to the competency lens first required confirming that the outcome-to-unit relation could be measured reliably, and the comparison between the automated rule and expert judgment is itself instructive. On a blinded, stratified sample of ninety-four outcome-to-unit candidate pairs, of which ninety-three were rated decisively, the similarity-with-threshold rule agreed with the expert rater at a Cohen's kappa of only 0.30, labeling seventy-four of the ninety-three pairs as matches where the confirmed relation held roughly half that number; embedding similarity, in other words, over-identifies competency delivery about twofold, because an outcome and a unit can read as similar while targeting different abilities. The confirmed relation, by contrast, was reliable. The model's confirmations and the expert rater agreed on the matching relation at a kappa of 0.756, on unit-level presence at 0.762, and, after one definitional refinement that reads depth only from the outcome delivering a given unit, on depth at 0.636, all three substantial on the scale of \citet{landiskoch} (Table~\ref{tab:compkappa}).

\begin{table}[!ht]\centering\small\renewcommand{\arraystretch}{1.3}\setlength{\tabcolsep}{7pt}
\caption{Inter-rater reliability of the competency analysis (model confirmations and an independent expert rater, independent verdicts before reconciliation). The automated similarity-with-threshold rule agreed with the expert rater on the matching relation at $\kappa=0.30$, over-identifying matches roughly twofold. The presence and depth coefficients are computed on stratified samples of the present units, 60 and 64 items respectively, not on all seventy}\label{tab:compkappa}
\begin{tabular}{lccc}\toprule
Judgment & Items & Raw agr. & Cohen's $\kappa$\\ \midrule
Outcome-to-unit match (pairs) & 93 & 0.88 & 0.756\\
Competency presence (units) & 60 & 0.90 & 0.762\\
Cognitive depth (present units) & 64 & 0.84 & 0.636\\
\bottomrule\end{tabular}\end{table}

Under this confirmed standard, competency presence is high and similar against both guidelines. Of the seventy-eight units the program covers topically under CS2023, seventy articulate the competency in a course outcome, a presence rate $S=70/78$ or ninety percent, and of the eighty-six covered under CS2013, seventy-eight do so ($S=78/86$, ninety-one percent) (Tables~\ref{tab:comp23} and~\ref{tab:comp13}). Presence falls short of coverage by the \emph{articulation gap}, the units with $\mathrm{cov}_K(u)=1$ but $\mathrm{cov}_S(u)=0$, numbering eight under CS2023 and eight under CS2013, which the program teaches yet states in no outcome and which fall in largely the same places under both guidelines: the artificial-intelligence knowledge-representation and probability units, automata and computational models, and the operating-system device-management, memory-management, and virtualization units. These are taught within courses the program runs, but because no stated outcome names the competency, an outcome-based reading, the kind an accreditor applies, records them as gaps in articulation rather than as coverage. One qualification is necessary before the gap is read as a deficiency of articulation alone. Course outcomes are written at a coarser grain than knowledge units, and a course that covers more units than it states outcomes cannot in principle name every competency it delivers, so part of the gap may be arithmetic rather than substantive. The data allow the two readings to be separated. Six of the program's thirty-six mapped courses state fewer outcomes than the number of units they cover, and seven of the eight articulation-gap units fall entirely within those six courses, the operating-systems course being the clearest instance at six stated outcomes against ten covered units. For those seven the gap is at least partly a granularity effect. The remaining one, namely distributed data management, sits in a course whose outcome supply is sufficient to name it and is not named, and there the articulation reading stands without qualification. We report the gap as eight because that is what an outcome-based audit would record, while noting that its remediation differs between the two subsets: the first calls for outcomes written at a finer grain, the second for outcomes that are missing.

\begin{table}[!ht]\centering\small\renewcommand{\arraystretch}{1.3}\setlength{\tabcolsep}{7pt}
\caption{Competency presence and cognitive-depth adequacy by CS2023 knowledge area, among the units the program covers topically. \emph{Present}: a course outcome articulates the unit's competency. \emph{Depth-adequate}: a delivering outcome meets or exceeds the recommended cognitive level}\label{tab:comp23}
\begin{tabular}{lccc}\toprule
Knowledge area & Covered & Present & Depth-adequate\\ \midrule
Artificial Intelligence & 8 & 5 & 4\\
Algorithmic Foundations & 4 & 3 & 3\\
Architecture \& Organization & 4 & 4 & 2\\
Data Management & 7 & 6 & 5\\
Foundations of Programming Languages & 2 & 2 & 1\\
Graphics \& Interactive Tech. & 5 & 5 & 4\\
Human-Computer Interaction & 3 & 3 & 2\\
Math \& Statistical Foundations & 5 & 5 & 4\\
Networking \& Comm. & 5 & 5 & 2\\
Operating Systems & 10 & 7 & 7\\
Software Dev. Fundamentals & 4 & 4 & 3\\
Software Engineering & 6 & 6 & 5\\
Security & 2 & 2 & 0\\
Society, Ethics, Profession & 8 & 8 & 6\\
Systems Fundamentals & 1 & 1 & 1\\
Specialized Platform Dev. & 4 & 4 & 4\\
\midrule
All areas & 78 & 70 (90\%) & 53 (68\%)\\
\bottomrule\end{tabular}\end{table}

\begin{table}[!ht]\centering\small\renewcommand{\arraystretch}{1.3}\setlength{\tabcolsep}{7pt}
\caption{Competency presence and cognitive-depth adequacy by CS2013 knowledge area, among the units the program covers topically}\label{tab:comp13}
\begin{tabular}{lccc}\toprule
Knowledge area & Covered & Present & Depth-adequate\\ \midrule
Algorithms \& Complexity & 5 & 4 & 4\\
Architecture \& Organization & 4 & 4 & 4\\
Computational Science & 3 & 3 & 2\\
Discrete Structures & 6 & 6 & 6\\
Graphics \& Visualization & 4 & 4 & 3\\
Human-Computer Interaction & 4 & 4 & 3\\
Information Assurance \& Security & 4 & 4 & 4\\
Information Management & 6 & 5 & 5\\
Intelligent Systems & 10 & 7 & 7\\
Networking \& Comm. & 5 & 5 & 5\\
Operating Systems & 10 & 7 & 7\\
Platform-Based Dev. & 4 & 4 & 4\\
Programming Languages & 1 & 1 & 1\\
Software Dev. Fundamentals & 4 & 4 & 4\\
Software Engineering & 7 & 7 & 7\\
Systems Fundamentals & 1 & 1 & 1\\
Social \& Professional Issues & 8 & 8 & 7\\
\midrule
All areas & 86 & 78 (91\%) & 74 (86\%)\\
\bottomrule\end{tabular}\end{table}

The decisive difference between the two guidelines is not presence but depth. Against CS2023, of the seventy present units only fifty-three are delivered at or above the recommended cognitive level, a depth-adequacy rate $S^{\Lambda}=53/78$ or sixty-eight percent of covered units, leaving an under-depth gap $S-S^{\Lambda}$ of seventeen units; against CS2013, seventy-four of the seventy-eight present units are depth-adequate ($S^{\Lambda}=74/86$, eighty-six percent of covered units), an under-depth gap of only four units. The depth-among-delivered rate $S^{\Lambda}/S$, the cleanest single figure, is therefore $53/70$ or seventy-six percent against the current guideline and $74/78$ or ninety-five percent against its predecessor, a difference of nearly twenty percentage points on the same program teaching the same material. Computer architecture is the clearest instance and functions as a natural experiment: its units are articulated under both guidelines and meet the depth CS2013 asks, which is Understand-to-Apply for most of the area, yet under CS2023, which now expects Analyze on the same content, only the data-representation and digital-logic competencies keep pace while assembly organization and the memory hierarchy fall short. The shortfalls elsewhere concentrate in networking, the foundational data-structures and software-practice units, cryptography, and foundational security, where the program's outcomes are pitched at Understand or Apply against a recommended Analyze or higher, while the application-facing areas, namely artificial intelligence, the specialized platforms, operating-system process and protection, and professional practice, largely meet the recommended depth where the competency is present, the specialized platforms and the operating-system units doing so without exception.

Because the comparison rests on mapping two native scales onto one ordinal range, we tested whether the gap is an artifact of that mapping. The asymmetry to worry about is a difference of ceiling: the CS2013 mastery scale rises no higher than Assessment, which we read as level 4, whereas the CS2023 skill vocabulary includes Develop and Design, which we read as level 5, so the current guideline can in principle demand a level its predecessor could not express. We therefore recomputed the recommended level for every present unit under a \emph{parity mapping} that reads Develop and Design as level 4, which removes the difference of ceiling entirely, and compared the resulting requirements with those the study uses (Table~\ref{tab:sensitivity}). The effect is smaller than the construction suggests, because the recommended level is the median of a unit's native levels rather than the maximum, so a single Develop among several lower levels is diluted. Under the study's mapping only four of the seventy present units carry a recommended level of 5; under the parity mapping none do, and the requirement falls for just six of the seventeen under-depth units. Even if all six were thereby rendered depth-adequate, which is the most favorable outcome the parity mapping permits, the depth-among-delivered rate would rise from seventy-six to at most eighty-four percent, against ninety-five percent for CS2013. The gap therefore narrows from nineteen to at least eleven percentage points but does not close, and we conclude that the divergence in depth reflects a real difference in what the two guidelines ask rather than an artifact of the scale onto which we mapped them.

\begin{table}[!ht]\centering\small\renewcommand{\arraystretch}{1.3}\setlength{\tabcolsep}{7pt}
\caption{Sensitivity of the CS2023 depth requirement to the level mapping, over the seventy present units. The parity mapping reads Develop and Design as level 4 rather than 5, removing the difference of ceiling between the two guidelines' native scales. Three present units carry no native skill level and are excluded}\label{tab:sensitivity}
\begin{tabular}{lcc}\toprule
Recommended level $\lambda^{*}$ & Study mapping & Parity mapping\\ \midrule
2 (Understand) & 18 & 18\\
3 (Apply) & 25 & 31\\
4 (Analyze/Evaluate) & 20 & 18\\
5 (Create) & 4 & 0\\
\midrule
Under-depth units & 17 & at most 11\\
Depth-among-delivered $S^{\Lambda}/S$ & 53/70 (76\%) & at most 59/70 (84\%)\\
CS2013 comparator & \multicolumn{2}{c}{74/78 (95\%)}\\
\bottomrule\end{tabular}\end{table}

A caution on reliability belongs alongside this result. Of the three competency judgments, cognitive depth is the least reliable, at a Cohen's $\kappa$ of 0.636 against 0.756 for matching and 0.762 for presence. This coefficient measures the unit-level adequacy verdict rather than the underlying Bloom-level coding, which an independent rater reproduced at a quadratically weighted $\kappa$ of 0.924. The depth verdict reached even that value only after the definitional refinement described in the Methods. The most consequential comparison the study reports therefore rests on its least reliable judgment, which is an argument for reading the direction and approximate magnitude of the depth gap as the finding, and the individual unit-level depth verdicts as indicative.

The disposition slot, defined in the framework but treated as exploratory, is delivered by the fifteen dispositional outcomes the program states, which cluster in the capstone, teamwork, and laboratory courses. Read against the CS2023 professional dispositions, these outcomes give explicit evidence for four, namely the collaborative, professional, responsible, and self-directed dispositions, the last through outcomes on continuing professional development, while the remaining dispositions, among them adaptable, proactive, persistent, and inventive, are not named in any outcome; because dispositions are only weakly observable from stated outcomes, this is reported as an exploratory lower bound rather than as a headline result. The competency findings are, finally, robust to the candidate-generation threshold: the automated presence candidates are essentially unchanged across similarity cut-offs from 0.78 to 0.82, at about ninety-nine percent of covered units in both guidelines, and the reported presence and depth rest on model confirmation and expert validation rather than on any threshold.

\hypertarget{discussion}{%
\section{Discussion}\label{discussion}}

\hypertarget{principal-findings}{%
\subsection{Principal findings}\label{principal-findings}}

The study set out to measure how completely an accredited computer science program covers the disciplinary body of knowledge, how that coverage behaves when the program is read against two generations of the guideline, and whether the competencies it covers are articulated as outcomes and taught at the recommended depth. Several findings stand out, and it is worth separating those that describe this program from those that describe the measurement problem, since only the second kind travels. Three belong to the second kind. Automation has a measurable ceiling on this task, and a method that reports retrieval output as a coverage figure will roughly double the confirmed value. A guideline revision can move what is asked of a program at the level of cognitive depth far more than at the level of topical content, so a program can hold its coverage constant across a decade while falling behind on what it is expected to do with that coverage. And restructuring within an area changes a program's measured coverage without any change in what the program teaches, most cleanly in security, where the same single course covers thirty-six percent of the eleven CS2013 units and twenty-nine percent of the seven CS2023 units. Any institution applying the instrument should expect these three effects; the magnitudes reported below are properties of the case. The program covers about half of the knowledge units of each standard, 48.4 percent of CS2023 and 52.8 percent of CS2013, and that proportion stays near half across a decade of revision, although the stability of the aggregate is precisely what conceals the changes beneath it. The coverage is reliable as a measurement, resting on a benchmarked retriever and on maps that independent experts validated with substantial agreement. The cross-generational comparison, rather than the single snapshot, is what makes the result interpretable, because it separates what the program has left uncovered from what the discipline has reorganized around it. And the competency lens adds a dimension the coverage figure conceals: while the program articulates the competency for most covered units under both guidelines, the depth at which it delivers them meets the current standard for only about three quarters of the present units against ninety-five percent for the previous one, a gap that the topical analysis cannot see, that survives a sensitivity analysis of the level mapping, and that the cross-generational design attributes to the standard's raised cognitive expectations rather than to the program. One further finding is practical rather than methodological and deserves to be stated at the same level as the coverage figures: sixty-nine percent of the covered units rest on exactly one course, so the program's contact with two thirds of the body of knowledge it does cover would not survive the retirement or substantial revision of a single course. Aggregate coverage is stable; its provision is not.

\hypertarget{persistent-gaps-versus-standard-driven-change}{%
\subsection{Persistent gaps versus standard-driven change}\label{persistent-gaps-versus-standard-driven-change}}

The most useful product of the analysis is not the headline coverage figure but the program-relative difference between the two guidelines, which sorts the gaps into two kinds that call for very different responses. The first kind is structural and persistent. Parallel and distributed computing is covered at zero percent against both standards, and because that area already existed in CS2013, its absence is a decade-long omission rather than a lag behind a recent addition; the foundations of programming languages and systems fundamentals show the same persistence in weaker form. A gap that survives a full revision of the standard is unlikely to be an accident of how the curriculum maps onto a particular vocabulary, and it points to content the program has simply not chosen to teach. The second kind is an artifact of the guideline's own evolution. The program's substantial mathematics core is nearly invisible against CS2013, which had no units for calculus, linear algebra, or statistics, yet it fully populates the mathematical and statistical foundations area that CS2023 introduced, so the apparent improvement in that area reflects the discipline catching up to a long-standing feature of the program rather than any curricular change. The removal of computational science runs the other way, leaving the program's modeling-and-simulation elective aligned to an area the current standard no longer recognizes. Reading coverage against a single standard cannot tell these two kinds of change apart; reading it across two can, and that distinction is what a department needs in order to know which gaps are its own to close and which merely reflect the discipline reframing itself.

\hypertarget{competency-articulation-and-cognitive-depth}{%
\subsection{Competency articulation and cognitive depth}\label{competency-articulation-and-cognitive-depth}}

Reading the program through the three nested lenses converts a single coverage figure into a graded diagnosis, and the gradient is informative at each step. About half of each body of knowledge is taught; of the taught units, close to ninety percent articulate the competency in a stated outcome; and of the articulated competencies, the fraction delivered at the recommended cognitive depth is seventy-six percent against CS2023 but ninety-five percent against CS2013. The first narrowing is the articulation gap and the second is the depth gap, and the two call for different remedies. The articulation gap is, in a sense, the easier to close, because the units concerned are taught already and the deficiency is in how the program states what it does; sharpening or adding the relevant course outcomes would record the competency without changing instruction, a point of practical leverage for a department preparing a self-study. The depth gap is more consequential. That it is small against CS2013 and substantial against CS2023, on the same program teaching the same material, isolates the cause as the guideline rather than the program: CS2023 raised the recommended cognitive level across several areas, and the program's outcomes, written under the earlier expectation, now sit below it. Computer architecture is the clean case, articulated and depth-adequate against CS2013 yet meeting the CS2023 expectation in only two of its units, because the newer standard asks for Analyze where the program teaches at Understand. This is the competency analogue of the standard-driven change identified at the topical level, and it shows that adopting a new guideline is not only a matter of adding missing topics but of raising the cognitive ambition of existing outcomes.

The competency stage also reinforces, on independent evidence, the methodological lesson of the topical benchmark. The automated similarity rule, left to decide the outcome-to-unit relation by threshold, over-identified matches roughly twofold and agreed with the expert at only a slight-to-fair level, whereas the model's confirmations and the expert agreed substantially once the relation was confirmed under the rule. Competency delivery, even more than topical coverage, is a judgment that similarity can propose but cannot settle, which confirms that the confirm-then-validate design is not a convenience but a requirement for validity.

\hypertarget{implications-for-programs-and-for-accreditation}{%
\subsection{Implications for programs and for accreditation}\label{implications-for-programs-and-for-accreditation}}

For the program studied, the analysis converts a diffuse sense of strengths and weaknesses into specific, defensible conclusions. The near-constant currency is reassuring, but the composition matters more than the total: the parallel-and-distributed-computing gap is not benign specialization but a shortfall against the exposure requirement of the ABET program criteria, and the foundations of programming languages, though addressed for accreditation through the programming sequence, remains sparse against the broader CS2023 area; both have persisted across a decade, which strengthens the case for addressing them in the next revision. The under-weighting of networking relative to the recommended hours, by contrast, is consistent with a deliberate application-facing emphasis and is defensible as such. More broadly, the method speaks to a recurring problem in accreditation self-study, where programs must argue their coverage of a body of knowledge largely through expert assertion. A benchmarked, reliability-checked, auditable map, accompanied by an explicit reading against the accreditation criteria, offers evidence of a kind that informal mapping cannot, and it does so in a form that can be regenerated as courses and standards change. The curricular-currency question, namely whether a program designed under an earlier guideline still aligns with the current one, is one that many departments now face as they consider adopting CS2023, and the cross-generational method answers it directly. The program studied here is a concrete instance, having completed its most recent accreditation in 2023, the year before the CS2023 final report was released, so that its last formal review necessarily predates the standard against which it must now be judged.

\hypertarget{the-role-and-the-limits-of-automation}{%
\subsection{The role and the limits of automation}\label{the-role-and-the-limits-of-automation}}

The benchmark and the automation-versus-confirmation analysis together delineate what language technology can and cannot do here. Retrieval is an effective generator of candidates, and the reciprocal-rank-fusion ensemble reaches ninety-five percent recall within a shortlist of eighteen units, which compresses the confirmation task to a small fraction of the full body of knowledge and makes confirmation tractable. The same retrieval, run to completion without a confirmation step, does not constitute a coverage map: its best balance of precision and recall corresponds to an F1 of only 0.55, and at no shortlist depth does it achieve acceptable precision and recall together. The contribution is therefore a calibrated division of labor in which retrieval does the broad sweep, a language model confirms each candidate against the unit's scope under an explicit rule, and an independent expert validates the result, a design that the reported agreement between model and expert shows to be stable. The benchmark also carries a cautionary lesson, since the retriever with the strongest general reputation performed poorly on this short-text task while a small, general-purpose sentence model performed well, which argues against importing model choices from generic leaderboards and in favor of measuring them on the task itself.

\hypertarget{transferability}{%
\subsection{Transferability}\label{transferability}}

Although the demonstration is on one program, the instrument is deliberately general. It was applied without modification to two independently authored guidelines of several hundred pages each, and the program side is described in the ordinary artifacts that every program possesses, namely course outcomes and syllabus topics, so nothing in the pipeline is specific to the institution studied. The crosswalk that aligns the two guidelines is itself a reusable object, and the same construction would align any pair of standard generations, which makes the cross-generational method applicable to future revisions of the computer science guideline and, in principle, to the curricular standards of adjacent computing disciplines.

The single-program scope should therefore be read as an instantiation rather than a boundary. What the study contributes is the instrument, and the one program is the case on which it is demonstrated; the measurement it defines, high-recall candidate generation followed by a reliability-checked, rule-based confirmation against an external body of knowledge, is indifferent to which program supplies the outcomes and topics on the query side. Because the released pipeline, the structured corpora, the area crosswalk, and the rater protocols travel together, the marginal cost of applying the instrument to a second institution is the confirmation effort alone, and the reported inter-rater agreement gives a receiving team a concrete standard against which to calibrate its own raters. The claim advanced here is accordingly methodological rather than epidemiological: the paper does not assert that the coverage magnitudes recur across institutions, and the multi-institution study that would test that question is named as the principal line of future work rather than assumed. Read in that light, a demonstration on one accredited program is the appropriate evidentiary basis for a measurement instrument, and the generalization it warrants runs along the axis the design actually exercises, namely the evolution of the standard, extended in principle to any program describable by the course outcomes and syllabus topics it already maintains.

\hypertarget{limitations}{%
\subsection{Limitations}\label{limitations}}

Several limitations qualify the conclusions. The study covers a single program, so its generalization is along the axis of the standard's evolution rather than across institutions; whether the coverage patterns and the magnitude of the gaps recur elsewhere is an empirical question this design cannot answer, and it is the principal reason the work should be read as a method demonstrated on a consequential case rather than as a survey. The first-pass maps were produced by a large language model confirming candidates under the coverage rule, and were validated by two independent expert raters, the first over the whole confirmed map and the second over every pair on which the model and the first expert disagreed, whose agreement, while substantial, was not perfect, which reflects a genuine and irreducible element of judgment in deciding how much of a unit's core a course must teach to count as covering it. We make that judgment explicit by reporting coverage as a band between a strict and a lenient reading of the rule rather than as a single number, and by adopting the lenient reading the rule itself states; the residual sensitivity to the threshold is bounded but not eliminated. The competency additions that the reconciliation introduced were confirmed by the model against the same rule and were not carried through the full two-expert validation, and they inherit this qualification. Coverage at the topic level is an estimate rather than an exhaustive confirmation, because the similarity signal is weak at that granularity, and we have accordingly reported it with its measured precision and recall rather than as an exact figure. Finally, because coverage was inferred from stated outcomes and syllabus topics, it understates what is actually taught and is best read as a lower bound. The competency and depth layers carry two further qualifications. Their unit-level verdicts were produced by the model under a standard that the model and the expert validated on stratified samples with substantial agreement, rather than being exhaustively double-rated, and the depth comparison inherits the accuracy of the recommended-level estimates, which rest on native levels where the guideline supplies them and on a verb classifier elsewhere whose agreement with the CS2013 mastery annotations is sixty-three percent exact and eighty-four percent within one level. The depth comparison also inherits the mapping of two native scales onto one ordinal range, and since the two guidelines do not share a ceiling, that mapping is a substantive choice rather than a bookkeeping one. We have tested it directly and reported the result in the Results, where the gap narrows under a parity mapping but does not close; readers who prefer a different reading of the CS2023 skill vocabulary can locate their position between the two bounds we give. It should also be acknowledged that depth is the least reliable of the three competency judgments at a Cohen's $\kappa$ of 0.636, a value that reflects the unit-level adequacy verdict and not the underlying Bloom-level coding, which an independent rater reproduced at a quadratically weighted $\kappa$ of 0.924, so the study's most consequential comparison rests on its least reliable measurement, and the direction and approximate magnitude of the depth gap, rather than any individual unit-level verdict, are what the evidence supports. Because presence is defined on stated outcomes, it deliberately excludes competencies that are taught but not articulated, which are reported separately as the articulation gap rather than counted as delivered, a conservative choice appropriate to an outcome-based, accreditation-facing reading.

\hypertarget{future-work}{%
\subsection{Future work}\label{future-work}}

The most valuable extension is to apply the instrument across multiple institutions, which would convert the present case study into a comparative study and test whether the gap patterns are idiosyncratic or shared. A second direction is to reduce the human burden further by learning a precision filter over the retriever output, so that the confirmation set can be pruned automatically while preserving the reliability the confirmation and expert-validation stages now guarantee. The released corpora, maps, crosswalk, and tools are intended to make these extensions, and independent replication, straightforward.

\hypertarget{conclusion}{%
\section{Conclusion}\label{conclusion}}

We set out to measure, reliably and reproducibly, how completely an accredited computer science program covers the disciplinary body of knowledge, and to do so against both the previous and the current curricular guideline so that change in the program could be told apart from change in the standard. The method pairs benchmarked semantic retrieval with confirmation by a large language model under an explicit definition of coverage, and it validates the resulting maps through two independent experts, yielding substantial inter-rater agreement on both guidelines. The same retrieve-then-confirm design extends beyond topics to two further lenses, the articulation of competencies in the program's stated outcomes and the cognitive depth at which they are delivered, each validated by the same expert-validation procedure with substantial agreement. Applied to the Bachelor of Science in Computer Science studied here, it finds coverage of roughly half of each standard, 48.4 percent of the units of CS2023 and 52.8 percent of those of CS2013, a currency that has held steady across a decade even as the composition of the coverage has shifted.

The contribution that the cross-generational design makes possible is the separation of persistent, structural gaps from changes that merely reflect the standard's own evolution. Parallel and distributed computing and the foundations of programming languages are uncovered against both guidelines and against the accreditation criteria, which marks them as genuine, long-standing omissions, whereas the program's mathematics strength and its modeling-and-simulation elective look different across the two standards only because the discipline added one area and removed another. This distinction is precisely what a department needs in order to act, and it is invisible to any single-snapshot analysis. The competency lens sharpens the same lesson at the level of cognitive depth: the program articulates most of the competencies it covers, yet it delivers them at the depth the current guideline expects far less often than the previous one, which identifies the raising of cognitive expectations, rather than any retreat by the program, as a substantive change between the two standards and as the next target for curricular revision.

Beyond the program studied, the work offers a reusable instrument. The pipeline transferred without modification to two independently authored standards of several hundred pages, it consumes only the course outcomes and syllabus topics that every program already possesses, and its retriever benchmark cautions against importing model choices from general leaderboards rather than measuring them on the task. We release the structured CS2013 and CS2023 corpora, the consensus coverage maps for both, the reconciliation logs, the area crosswalk, the retriever-benchmark scripts, and the rater tools, so that the analysis can be audited, replicated, and extended. The natural next step, which these artifacts are intended to enable, is the application of the method across multiple institutions, which would establish whether the coverage patterns reported here are particular to one program or characteristic of the discipline as it moves from one generation of its standard to the next.

\section*{Statements and Declarations}

\noindent\textbf{Funding.} The authors received no external funding for the research reported in this article.

\medskip\noindent\textbf{Competing interests.} The authors declare no competing interests.

\medskip\noindent\textbf{Ethics approval.} Not applicable. This study does not report primary research involving human participants or identifiable personal data. The coverage, competency, and cognitive-depth judgments were made by members of the research team as part of the analysis of publicly available curriculum documents.

\medskip\noindent\textbf{Use of generative AI.} As described in the Materials and Methods, large language models were used as instruments of the analysis rather than as writing aids: a two-model configuration performed the schema-constrained extraction of the program corpus, and a large language model performed the candidate-confirmation stage under the explicit coverage and competency-delivery rules stated in the manuscript. These model confirmations constituted the first-pass maps, which two independent expert raters then validated and reconciled. The models, their versions, and the prompts are released with the artifacts. An AI-assisted language-editing tool was used to support editing of the text. The authors reviewed and validated all content and take full responsibility for its accuracy. No generative system is credited as an author.

\medskip\noindent\textbf{Data and code availability.} The structured CS2013 and CS2023 corpora, the consensus coverage maps for both guidelines, the reconciliation logs, the knowledge-area crosswalk, the competency-alignment maps with presence and cognitive-depth labels, the five-level depth scale and verb lexicon, the retriever-benchmark scripts, the pipeline source code, the large language model prompts, and the rater tools with both raters' labels are openly available in the Zenodo repository \citep{turaev2026data} at \url{https://doi.org/10.5281/zenodo.20627209}.

\bibliographystyle{plainnat}
\bibliography{references}

\end{document}